\documentclass{article}

\usepackage{microtype}
\usepackage{graphicx}
\usepackage{subcaption}
\usepackage{booktabs} 

\usepackage{hyperref}
\usepackage{multirow}
\usepackage{makecell}   
\usepackage{booktabs}   
\usepackage{graphicx}   

\usepackage[accepted]{icml2026}

\usepackage{amsmath}
\usepackage{amssymb}
\usepackage{mathtools}
\usepackage{amsthm}
\usepackage{hyperref}

\usepackage[capitalize,noabbrev]{cleveref}

\theoremstyle{plain}

\theoremstyle{definition}

\theoremstyle{remark}

\usepackage[textsize=tiny]{todonotes}

\icmltitlerunning{Measurement Plasticity: Sensor-Level Adaptation for Vision–Language Models}

\usepackage{amsmath} 
\DeclareMathOperator*{\argmax}{arg\,max}

\begin{document}

\twocolumn[
  \icmltitle{Measurement Plasticity: Sensor-Level Adaptation for Vision–Language Models}



  \icmlsetsymbol{equal}{*}

  \begin{icmlauthorlist}
    \icmlauthor{Boyeong Im}{snu}
    \icmlauthor{Wooseok Lee}{snu}
    \icmlauthor{Yoojin Kwon}{snu}
    \icmlauthor{Hyung-Sin Kim}{snu}
  \end{icmlauthorlist}

  \icmlaffiliation{snu}{Graduate School of Data Science, Seoul National University}

  \icmlcorrespondingauthor{Hyung-Sin Kim}{hyungkim@snu.ac.kr}

  \icmlkeywords{Machine Learning, ICML}

  \vskip 0.3in
]



\printAffiliationsAndNotice{}  

\newcommand{\ours}{\textit{MVP}}

\begin{abstract}

We propose Multi-View Physical-prompt (\ours) for Test‑Time Adaptation (TTA), a forward-only framework that moves TTA from tokens to photons by treating the camera exposure triangle (i.e., ISO, shutter speed, and aperture) as physical prompts.
At inference, \ours{} acquires selected multiple physical views using a source‑affinity score, evaluates digitally augmented variants of each retained view and filters the lowest-entropy predictions, and aggregates predictions with hard voting.
This selection‑then‑vote design is simple, calibration‑friendly, and requires no gradients or model modifications. 
On ImageNet‑ES and ImageNet‑ES‑Diverse, \ours{} outperforms digital‑only TTA on both Auto‑Exposure and a combination with conventional sensor control.
\ours~remains effective under reduced parameter candidates that lower capture latency, demonstrating its practicality. 

\vspace{-2ex}
\end{abstract}
\section{Introduction}

Foundation models (FMs) are increasingly deployed where inputs differ from the static web corpora on which they are trained, motivating training-free continual adaptation to changing inputs. 
Most existing approaches perform adaptation inside the model via weights updates, adapters, prompt tuning, or retrieval.
~\cite{wangtent, niutowards, hulora, shu2022test, karmanov2024efficient}.
In sensor-mediated environments, however, there is another natural locus of plasticity: the measurement process.
Vision–language models (VLMs) deployed in the physical world do not receive images from the web. 
Their inputs are produced by sensors, and camera settings such as ISO, shutter speed, and aperture determine what information reaches the encoder. 
When a scene is under-exposed, saturated, or noisy, downstream adaptation can only operate on an already degraded measurement. 
This suggests a complementary question for continual adaptation: instead of adapting the model to the input, can we adapt how the input is measured?

\begin{figure}
    \centering
    \includegraphics[width=0.9\linewidth]{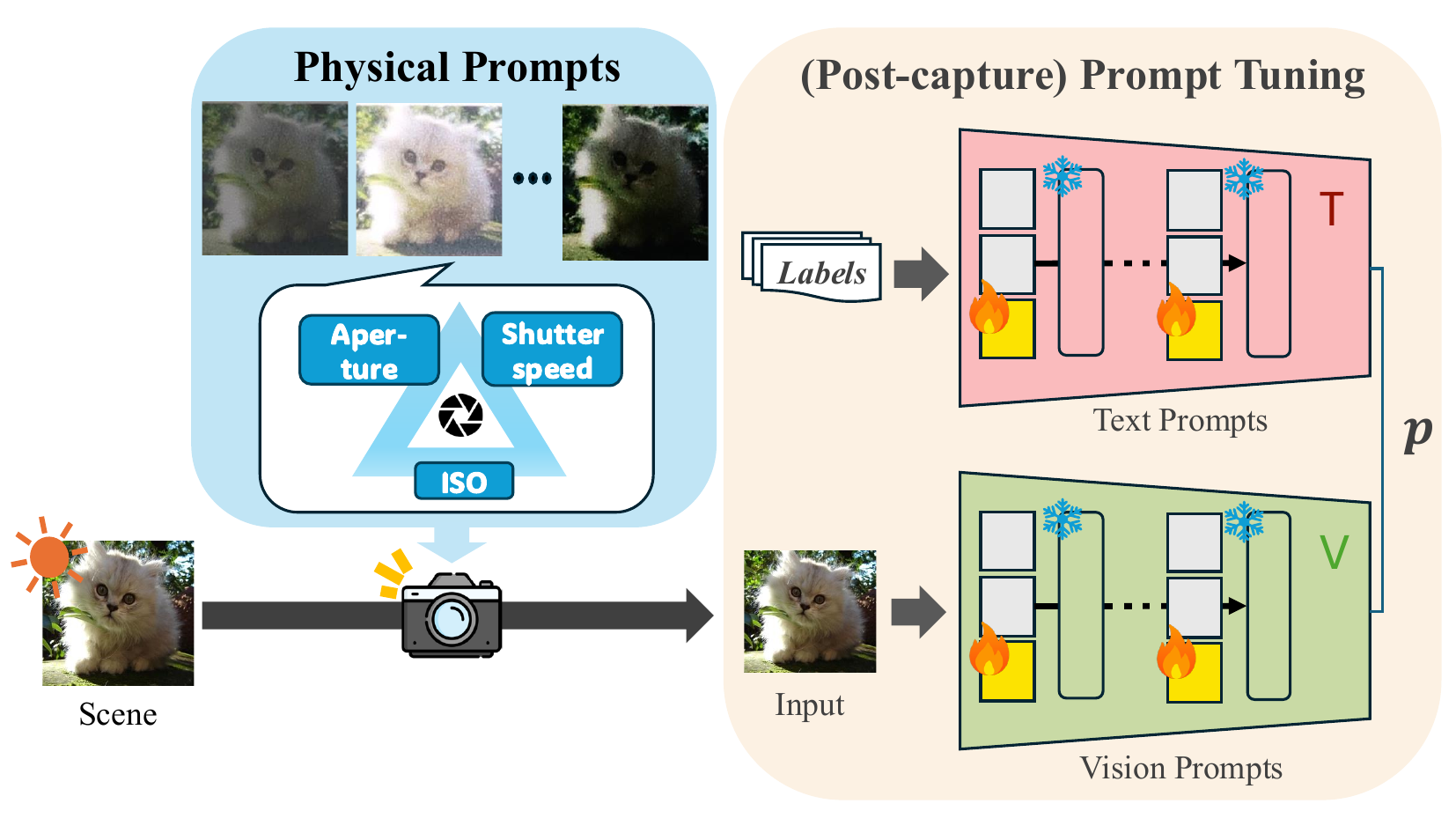}
    \vspace{-2ex}
    \caption{\textbf{Conceptual Figure of the Physical Prompts.} In the causal chain of scene $\to$ measurement $\to$ representation, physical prompts are controlled by sensor parameters (i.e., ISO, shutter speed) to minimize irrecoverable loss during measurement, which cannot be compensated from post-capture images.}
    \label{fig:conceptual}
    \vspace{-4ex}
\end{figure}

We study this question through measurement-level adaptation for VLMs. 
Our framework, Multi-View Physical-prompt for Test-Time Adaptation (\ours), treats the camera exposure triangle as a set of physical prompts (\cref{fig:conceptual}). 
At inference time, \ours~acquires multiple physical views of a scene under different sensor settings, selects views whose image features are closest to source-domain statistics, filters uncertain augmented views, and aggregates predictions using zero-temperature hard voting. 
\ours~is forward-only and requires no gradient updates or model modifications, making it a lightweight alternative to model-side adaptation.

This perspective reframes physical prompting as input-side continual adaptation. Instead of changing the VLM’s parameters, \ours~changes the distribution of measurements presented to the frozen model, shifting inputs closer to the regime where their representations are reliable. 
In practice, this multi-view framework is uniquely suited for static, precision-critical settings, such as CCTV monitoring, autonomous inspection, and computer-assisted surgery.
Experiments on ImageNet-ES~\cite{baek2024unexplored} and ImageNet-ES-Diverse~\cite{baek2025adaptive} show that \ours~improves robustness over digital-only Test-Time Adaptation (TTA) on Auto-Exposure captures and over pipelines that combine conventional sensor control~\cite{baek2025adaptive} with TTA, while remaining effective under reduced candidate sets that lower capture latency. 
These results suggest that sustainable adaptation for multi-modal FMs should consider not only where plasticity resides inside the model, but also whether plasticity can occur at the sensor–model interface.

Our contribution is to identify measurement plasticity as a first-class mechanism for continual adaptation in sensor-mediated AI. 
\ours~motivates future adaptation methods that operate jointly over models, prompts, memories, and the controllable processes that produce their inputs.
\section{Related Work}

\paragraph{Post-capture VLM adaptation.}
VLMs~\cite{radford2021learning, jia2021scaling} achieve strong zero-shot performance via large-scale vision--language pre-training. 
To improve downstream transfer, prompt tuning methods learn prompts in the language branch~\cite{zhou2022learning, zhou2022conditional} or across both modalities~\cite{khattak2023maple}, but require annotated downstream data. 
Test-time prompt tuning removes this requirement by adapting prompts per test sample~\cite{shu2022test}, with calibration-aware variants~\cite{yoon2024c, sharifdeen2025tpt} and memory-based extensions~\cite{ma2023swapprompt, zhang2024historical, xiao2025dynaprompt}.
PromptAlign~\cite{abdul2023align} aligns test visual-token statistics with source-domain statistics, which motivates our use of source affinity for view selection. 
Forward-only TTA methods reduce test-time cost by avoiding backpropagation
\cite{zanella2024test, sui2025just, farina2024frustratingly, karmanov2024efficient}. 
However, these methods still operate after capture: they adapt prompts, features, caches, or predictions for a fixed physical measurement.
\vspace{-2.5ex}
\paragraph{Sensor-level shift and adaptation.}
Sensor-level covariate shifts arise from the interaction between lighting and camera parameters (i.e., ISO, shutter speed). 
ImageNet-ES~\cite{baek2024unexplored} and ImageNet-ES-Diverse~\cite{baek2025adaptive} expose shifts via controlled light--sensor variations for static scenes, showing robustness gaps that digital-only adaptation cannot fully close. 
Lens~\cite{baek2025adaptive} addresses this issue by changing capture, using a post-hoc, camera-agnostic controller that selects sensor settings per scene based on model confidence. 
Yet confidence-only single-view selection is prone to overconfident misclassification, leaving open VLM adaptation to sensor-driven input distribution.
\ours~addresses this gap through measurement plasticity: instead of only adapting the model after an image is captured, it adapts the sensor--model interface that determines what visual evidence reaches the VLM.

\section{Method}

\begin{figure}[t]
    \centering
    \includegraphics[width=\linewidth]{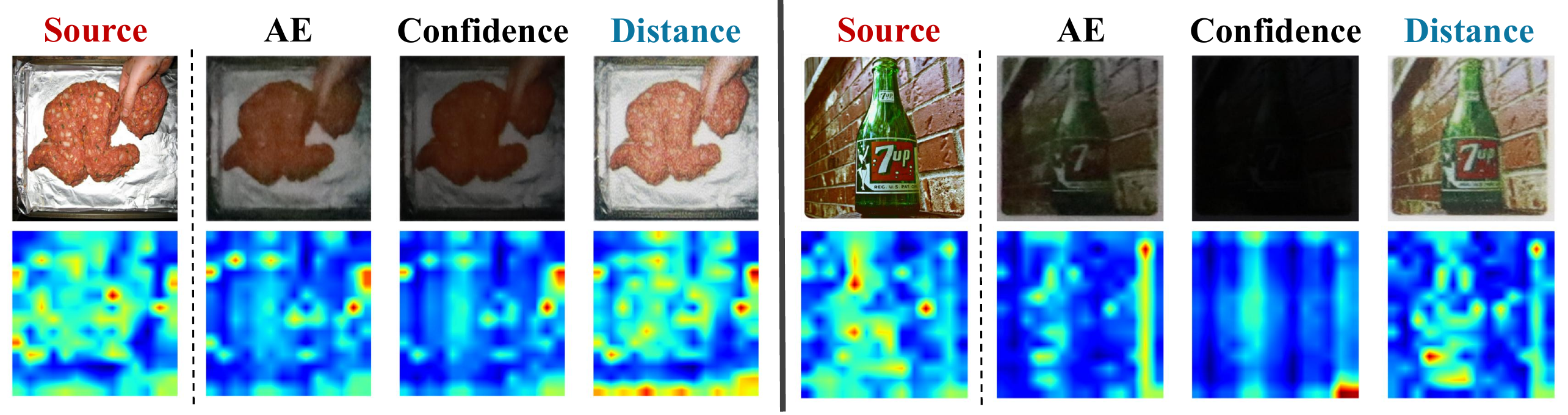} 
    \vspace{-4ex}
    \caption{\textbf{Visualization of various camera parameters.} From the identical scene, we compare the attention maps of the source (ImageNet) and varying parameter samples (ImageNet-ES-Diverse). 
    }
    \label{fig:att}
    \vspace{-4ex}
\end{figure}

\begin{figure*}[t]
    \centering
    \includegraphics[width=.9\linewidth]{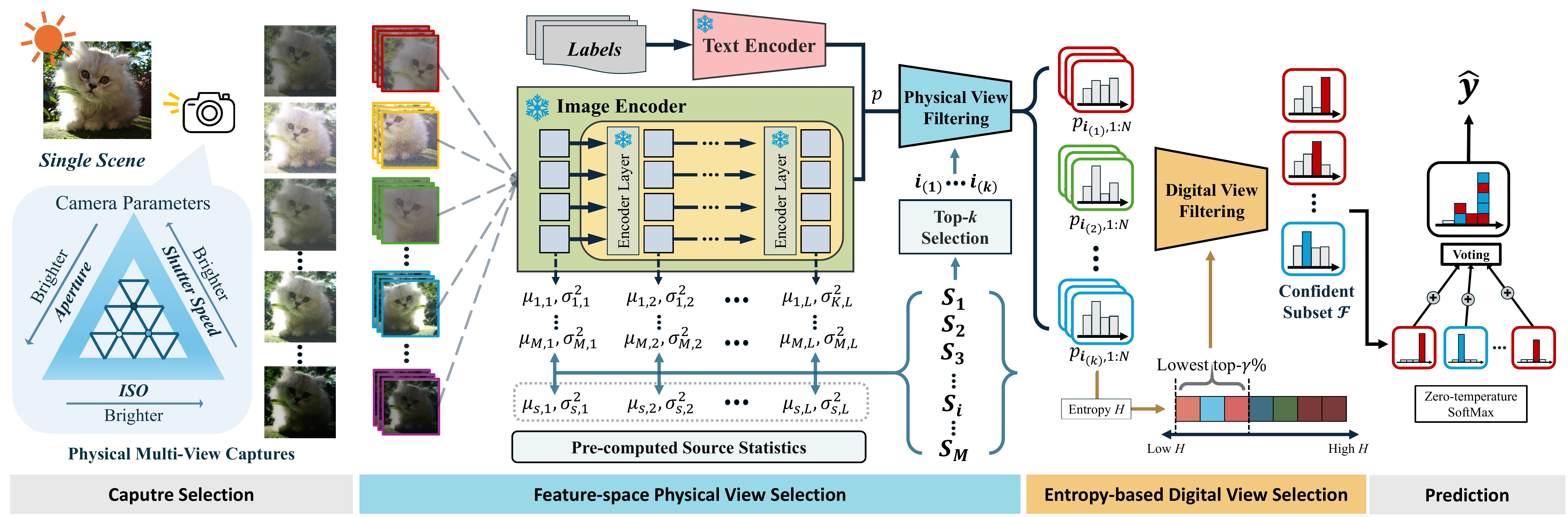}    
    \vspace{-1ex}
    \caption{\textbf{Overall Framework.} Given a single scene, multiple physical captures are obtained by varying camera parameters along the exposure triangle—ISO, shutter speed, and aperture—forming a set of physical multi-views, as controllable physical prompts.
    }
    \label{fig:overview}
    \vspace{-3ex}
\end{figure*}

Modern TTA methods for VLMs operate \emph{after capture}, adjusting how the model interprets a fixed image. With Auto-Exposure (AE), however, the captured scene may already have lost information (e.g., saturation or low photon count). Thus, digital TTA faces a hard practical bound: once photons are not measured, downstream adaptation often cannot recover them. 
This motivates a \emph{sensor-aware} view of TTA that respects the causal chain: scene →  measurement → representation.
We therefore treat camera sensor parameters
as \emph{physical prompts}: upstream control variables that determine which photons reach the sensor.

\subsection{Source-Affinity Selection}
\vspace{-1ex}
Confidence-based sensor control~\cite{baek2025adaptive} can be brittle: a shifted capture may produce high confidence while inducing unreliable VLM features. 
We instead select measurement settings by source affinity. 
Following PromptAlign~\cite{abdul2023align}, \ours~scores each candidate capture by the distance between its visual-token statistics and pre-computed source-domain statistics. 
Figure~\ref{fig:att} visualizes the intuition: sensor parameters change what the model attends to, and source-affinity-selected views yield attention maps closer to the source than AE or confidence-selected captures. 
Intuitively, source-affinity-selected views let the model ``see'' the scene through source-like visual evidence.
Supporting diagnostics on feature distance and attention similarity are included in Appendix~\ref{appendix:motivation}.

\subsection{Multi-view Physical-prompt for TTA}
\vspace{-1ex}
\paragraph{Selecting physical views in feature-space.}
Figure~\ref{fig:overview} shows the overall pipeline of \ours. Instead of AE capture in the conventional work, 
\ours~obtains the most source-aligned views from multiple $M$ captures of a single scene under different camera settings.
Each physical capture $v_i\in \mathcal{V} = \{v_1, v_2, ..., v_M\}$ is expanded into a set of $N$ digital augmentations $\{v_{i,1}, ..., v_{i,N}\}$, including the original.
For each capture $v_i$, we derive $N^{\prime}$ confident augmentations by selecting the top $\alpha$ fraction of $N$ total augmentations based on confidence. Then, for each layer $l$, we extract the feature-level mean and variance of image token embedding $\mu_{i,l} = \frac{1}{N^{\prime}} \sum_{n=1}^{N^{\prime}}\mu_{i,l,n} \ ;\ \sigma^2_{i,l} = \frac{1}{N^{\prime}} \sum_{n=1}^{N^{\prime}}\sigma^2_{i,l,n}$ from the frozen visual encoder using $N^{\prime}$ views.
These statistics are then compared against the pre-computed source-domain statistics $(\mu_{s,l}, \sigma_{s,l}^{2})$. 
%
Since CLIP's pretraining data is non-public, we use ImageNet as a proxy source dataset and additionally evaluate LAION-based statistics (Appendix~\ref{appendix:proxy}).
To measure the similarity of physical views with source image in feature-space, a source-affinity score is defined as:
\vspace{-1ex}
\begin{equation}
    S_i = -\frac{1}{L} \sum_{l=1}^{L} 
    \left( \| \mu_{i,l} - \mu_{s,l} \|_2^2 + \| \sigma_{i,l}^2 - \sigma_{s,l}^2 \|_2^2 \right),
    \label{eq:affinity_score}
\end{equation}
where $L$ denotes the number of layers of the visual encoder.
We select the top-$k$ settings maximizing $S_i$, corresponding to most source-aligned physical views. This replaces prompt optimization with gradient-free \emph{physical-prompt selection}, preserving gray-box compatibility and low computations.

\paragraph{Entropy-based filtering for augmented views.}
Even among the selected top-$k$ physical parameters, some augmented samples may produce uncertain or noisy predictions due to local variations in illumination or visual context. To mitigate the negative impact of unreliable samples, we measure how uncertain the model is to each augmented view using entropy $H_{i,n} = - \sum_{c} p_{i,n}(c) \log p_{i,n}(c)$,
where $p_{i,n}(c)$ denotes the predicted probability of class $c$ for the $n$-th digital augmentation of the $i$-th physical capture. Across all $k\times N$ augmented views obtained from the selected $k$ physical captures, we retain only the bottom $\gamma$\% in entropy as the most confident subset $\mathcal{F}$. 
The predictions of these entropy-filtered views are then aggregated using
a hard voting scheme across all selected views:
\vspace{-1.5ex}
\begin{equation}
    \hat{y} = \argmax_{y \in \mathcal{C}} \sum_{(i,n) \in \mathcal{F}}
    \mathbf{1}\left[\argmax_{c \in \mathcal{C}} p_{i,n}(c) = y\right],
\end{equation}
where $\mathcal{C}$ is the set of class labels. This aggregation mitigates the overconfidence issues commonly observed in averaging softmax probabilities across multi-views, while preserving the robustness from physical multi-view observations.

\paragraph{Physical multiviews as augmentations.}
Rather than collapsing sensor control to a single capture, \ours~votes over the top-$k$ source-affinity-selected physical views. 
Because ISO, shutter speed, and aperture change the measurement itself, these captures provide a physical augmentation axis that post-capture crops, flips, or photometric perturbations~\cite{buslaev2020albumentations} cannot fully emulate. 
This introduces useful degrees of freedom for observing the same scene through meaningfully different measurements and yields more robust hard-voting predictions than single-setting inference.
LDA diagnostics supporting this distinction are provided in Appendix~\ref{appendix:physical_aug}.

\section{Experiments}

\subsection{Experimental Setup}
\vspace{-1ex}
For each lighting condition, every sample is photographed for 5 times using the AE setting, forming the \textit{auto-exposure set}, and captured with 27 controlled combinations of sensor parameters—three levels each of ISO, aperture, and shutter speed—forming the \textit{parameter-controlled} set.
\vspace{-2ex}

\paragraph{Datasets and baselines.}
We evaluate our method on sensor-level shift benchmarks, 
ImageNet-ES~\cite{baek2024unexplored} and ImageNet-ES-Diverse~\cite{baek2025adaptive}, 
both constructed from Tiny-ImageNet~\cite{le2015tiny}. 
We compare \ours~to representative approaches across three categories; prompt learning methods,
prompt-based TTA methods, 
and training-free TTA methods.
We also include Lens~\cite{baek2025adaptive} as the sensor-control baseline, which selects the most confident physical capture per scene.
Further details are provided in the Appendix \ref{appendix:dataset} and \ref{appendix:implementation}.
\vspace{-1ex}

\paragraph{Evaluation protocols.}
\label{para:eval_protocol}
For each scene, we evaluate all baselines under three sensor-control settings:
(1) Auto-Exposure (AE): Without any sensor control, the results are obtained by averaging all five AE captures.
(2) AE with digital photometric augmentation: For the same AE captures, additional photometric augmentations (\textit{i.e.,} hue, saturation, brightness)~\cite{buslaev2020albumentations} are incorporated into the augmentation process within each TTA pipeline to simulate diverse exposure variations.
(3) Lens-based selection: following the procedure in~\cite{baek2025adaptive}, we first identify the most confident physical view among 27 candidates using CLIP-based MaPLe inference, then apply each baseline method to the selected view.


\begin{table}[t]
    \centering
    \setlength{\tabcolsep}{6pt}
    \caption{\textbf{Performance comparison of vision-language adaptation methods.}
    All experiments are conducted using the ViT-B/16 backbone. 
    For \textit{with Lens}, all tuning-based methods are applied after sensor control via Lens~\cite{baek2025adaptive}. 
    Results are reported by average across light environments and detailed per-light environment results are provided in the Appendix.
    }
    \vspace{-1ex}
        \resizebox{\linewidth}{!}{
        \begin{tabular}{c|c|c|ccc|ccc}
            \toprule
            & \multirow{3}{*}{\makecell{\textbf{Method}}}
            & \multirow{3}{*}{\makecell{\textbf{Tiny-} \\ \textbf{ImageNet}}}
            & \multicolumn{3}{c|}{\textbf{ImageNet-ES}}
            & \multicolumn{3}{c}{\textbf{ImageNet-ES-Diverse}} \\
            \cmidrule(lr){4-6}
            \cmidrule(lr){7-9}
            & & & \multirow{2}{*}{AE} & \multirow[c]{2}{*}{\makecell{AE + \\ Photo Aug.}} & \multirow[c]{2}{*}{\makecell{with \\ Lens}} 
            & \multirow{2}{*}{AE} & \multirow[c]{2}{*}{\makecell{AE + \\ Photo Aug.}} & \multirow[c]{2}{*}{\makecell{with \\ Lens}} \\
            & & & & & & & & \\
            \midrule
            
            \multirow{1}{*}{\makecell{{Pretrained}}}
            & CLIP       & 87.6  & 48.98 & -     & 82.45 & 37.65 & -     & 61.13 \\
            \midrule
            \multirow{2}{*}{\makecell{Prompt \\ Learning}}
            & CoOp          & 89.2  & 51.82 & -     & 84.90 & 40.47 & -     & 65.15 \\
            & MaPLe         & 89.0  & 52.57 & -     & 84.40 & 39.47 & -     & 64.68 \\
            \midrule
            \multirow{4}{*}{\makecell{Prompt-\\based \\ TTA}}
            & TPT                & 89.5  & 55.66 & 49.66 & 84.45 & 41.20 & 40.82 & 64.90 \\
            & PromptAlign     & 90.0  & 55.45 & 50.27 & 84.75 & 41.51 & 41.04 & 64.90 \\
            & C-TPT                & 90.5  & 53.55 & 45.92 & 84.55 & 39.88 & 39.09 & 64.72 \\
            & O-TPT        & 89.3  & 52.70 & 44.56 & 84.50 & 39.14 & 38.47 & 64.63 \\
            \midrule
            \multirow{3}{*}{\makecell{Training\\free \\ TTA}}
            & MTA            & 90.6  & 56.56 & 54.64 & 84.70 & 41.70 & 41.33 & 65.08 \\
            & TDA      & 90.0  & 58.17 & 57.29 & 84.80 & 40.78 & 40.83 & 64.85 \\
            & ZERO   & 90.6  & 57.05 & 56.77 & 84.80 & 39.91 & 39.80 & 64.67 \\
            \midrule
            \multicolumn{2}{c|}{\textbf{\ours~(ours)}} & - & \multicolumn{3}{c|}{\textbf{87.85}} & \multicolumn{3}{c}{\textbf{67.28}} \\
            \bottomrule
        \end{tabular}
        }
    
    \label{tab:main_vit_maple}
    \vspace{-2ex}
\end{table}

\subsection{Results}
\vspace{-1ex}
\paragraph{Main evaluation.}
We investigate the effectiveness of \ours{} against post-capture TTA baselines. Table~\ref{tab:main_vit_maple} summarizes the results across three sensor-control protocols.
%
Under AE, the zero-shot CLIP performance degrades significantly across lighting conditions, 
with existing TTAs offering only marginal gains. This results indicate that post-capture adaptation alone cannot recover information loss at sensor-level. 
Integrating digital photometric augmentations, which simulate various lightings and exposures, also proved ineffective at improving performance, confirming that digital perturbations cannot fully reproduce the physical effects of sensor changes.
%
In contrast, \textbf{\ours~outperforms all AE-based TTAs}, achieving significant gains of at least \textbf{29.68 pp} and \textbf{25.58 pp} on ImageNet-ES and ImageNet-ES-Diverse, respectively. This highlights the critical need for TTA strategies that explicitly incorporate sensor-level diversity at capture time. 
Our method also yields a robust performance gain of up to \textbf{3.4 pp} over pipelines combining Lens with TTA methods, validating the efficacy of multiple physical parameters.
These findings emphasize that \ours~ mitigates the sensitivity to suboptimal sensor-setting and enhances robustness against sensor-level shifts.

\vspace{-2ex}
\paragraph{Evaluation with limited parameter settings.}
While the number of combinations of possible camera parameter grows exponentially, capturing all camera configurations is infeasible given the high cost of acquisition.
To reduce the redundancy of physical capturing, we further utilize Candidate Selection Algorithms (CSA)~\cite{baek2025adaptive} as a pre-selection strategy that reduces the candidate sensor parameter space to a smaller set of $M$ discrete grids.
%
As shown in Table~\ref{tab:csa_results}, \ours~consistently outperforms baselines across different CSA. This demonstrates robustness and scalability under reduced-capture regimes, thereby supporting its practical use in real-world deployments where capture cost is non-negligible.
Details on the CSA and full results of different $M$ are in Appendix \ref{appendix:csa} and \cref{tab:appendix_csa1,tab:appendix_csa2,tab:appendix_csa3}.

\begin{table}[t]
    \centering
    \footnotesize
    \caption{\textbf{Performance Across Different Capture Scenarios (CSA).}
    Without CSA, the capture latency is the full ImageNet-ES setting -- 2.41 seconds per scene. 
    For CSA1 / CSA2 / CSA3, the number of captures, $M$, are 12 / 6 / 21, respectively. 
    }
    \label{tab:csa_results}
    \setlength{\tabcolsep}{1mm} 
    \resizebox{0.8\linewidth}{!}{
    \begin{tabular}{c|c|c|c|c|c|c}
        \toprule
        \multirow{4}{*}{\textbf{Method}} & \multicolumn{2}{c|}{\makecell{\textbf{CSA 1} \\ \textbf{(1.06 sec)}}} & \multicolumn{2}{c|}{\makecell{\textbf{CSA 2} \\ \textbf{(0.37 sec)}}} & \multicolumn{2}{c}{\makecell{\textbf{CSA 3} \\ \textbf{(0.91 sec)}}} \\
        \cline{2-7} 
        & \multirow[c]{2}{*}{\scriptsize\textbf{IN-ES}} 
        & \multirow[c]{2}{*}{\scriptsize\textbf{\makecell{IN-ES\\-Diverse}}}
        & \multirow[c]{2}{*}{\scriptsize\textbf{IN-ES}}
        & \multirow[c]{2}{*}{\scriptsize\textbf{\makecell{IN-ES\\-Diverse}}}
        & \multirow[c]{2}{*}{\scriptsize\textbf{IN-ES}}
        & \multirow[c]{2}{*}{\scriptsize\textbf{\makecell{IN-ES\\-Diverse}}} \\
        & & & & & & \\
        \midrule
        Lens & 84.75 & 61.79 & 84.38 & 61.76 & 85.00 & 54.38 \\
        \midrule
        + TPT & 84.88 & 62.22 & 84.55 & 62.49 & 85.05 & 62.29 \\
        + PromptAlign & 84.95 & 62.03 & 84.32 & 62.56 & 85.25 & 62.37 \\
        + C-TPT & 84.85 & 61.90 & 84.68 & 61.83 & 85.10 & 61.91 \\
        + O-TPT & 84.77 & 61.66 & 84.42 & 61.27 & 85.03 & 61.77 \\
        + MTA & 85.37 & 62.44 & 85.10 & 62.74 & 85.33 & 62.62 \\
        + TDA & 85.30 & 61.71 & 84.70 & 62.08 & 85.28 & 62.33 \\
        + ZERO & 85.43 & 61.49 & 84.60 & 61.96 & 85.62 & 61.69 \\
        \midrule
        \textbf{\ours~(ours)} & \textbf{87.27} & \textbf{63.79} & \textbf{86.65} & \textbf{63.79} & \textbf{87.87} & \textbf{64.41} \\
        \bottomrule
    \end{tabular}}
    \vspace{-1ex}
\end{table}

\begin{figure}
    \centering
    \includegraphics[width=0.85\linewidth]{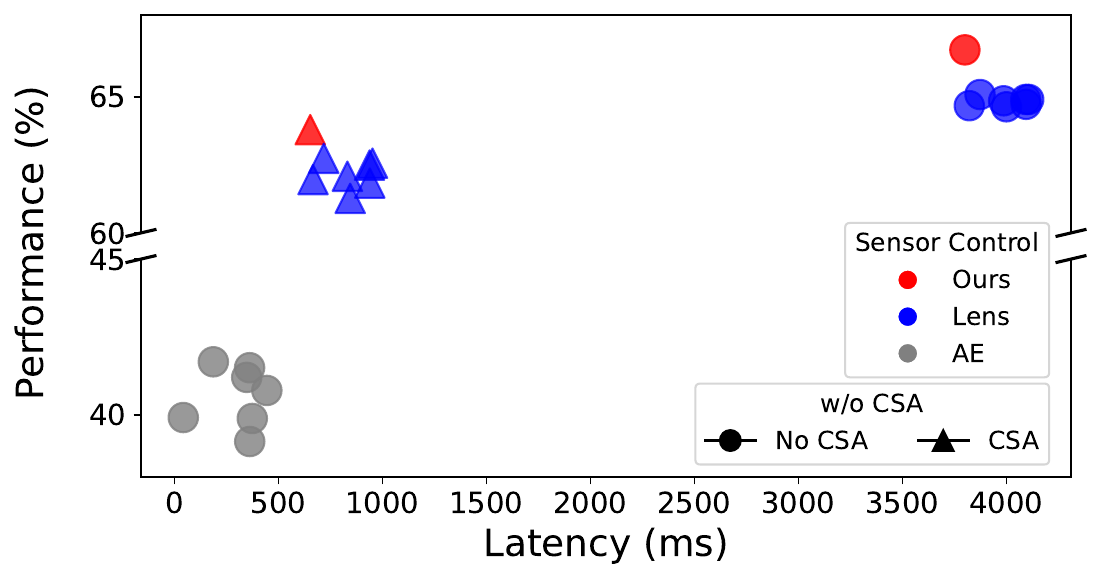}
    \vspace{-2ex}
    \caption{\textbf{Latency (ms) of adaptation and inference.} Experiments were conducted on NVIDIA RTX 4090 GPU and Intel(R) Xeon(R) Gold 6326 CPU. Color denotes the sensor control strategy, and markers distinguish TTA variant under the same strategy.}
    \label{fig:latency}
    \vspace{-2ex}
\end{figure}

\vspace{-1ex}
\paragraph{Runtime latency.}
\label{sssec:latency}
Figure~\ref{fig:latency} illustrates that \ours~delivers a more effective trade-off between performance and runtime latency (ms)---including capture time, adaptation, and inference--- per sample compared to baselines.
Specifically, AE attains the lowest latency but the weakest accuracy, consistent with its optimization for human perception rather than model compatibility. Without CSA ($M=27$ physical views), both \ours~and Lens exhibit the expected performance–latency trade‑off relative to AE, delivering $\ge$23.3 pp higher top‑1 accuracy at roughly 4$\times$ latency, reflecting the cost of capturing 27 physical views per scene. 
Notably, when CSA is applied ($M=6$), \ours~(red triangle) maintains high performance even at low latency which is similar to AE settings.
%
Compared to Lens-applied TTA settings (blue markers), \ours~achieves the highest accuracy at lower latency than competing baselines, demonstrating its strong efficiency--adaptability balance.
We also include qualitative comparisons that visualize the final selected views of MVP and ZERO with Lens, and ablations on MVP’s design choices, in Appendix \ref{appendix:qual} and \ref{appendix:ablation}.


\section{Conclusion}

In this paper, we introduce \ours, a novel framework that integrates sensor control into the TTA paradigm by interpreting the camera setting
as a \textit{physical prompt} for VLMs. 
By treating the sensor as a controllable prompt, our approach enables the model to decide \textit{what to capture}, extending conventional post-capture TTA into the sensor domain. 
Experiments show that our approach consistently outperforms baselines with high efficiency, indicating that robust real-world VLM generalization requires adapting not only how to interpret visual inputs, but also what to measure.


\bibliography{reference}
\bibliographystyle{icml2026}

\newpage
\appendix
\onecolumn

\section*{\huge Appendix}

\vspace{1em}

\section{Additional Motivation for Source-Affinity Selection}
\label{appendix:motivation}

We provide additional diagnostic evidence supporting the use of source-affinity selection in MVP.
These analyses are used only for motivation and do not require labels or attention maps at inference time.

\begin{figure}[h]
    \centering
    \begin{subfigure}[t]{0.35\linewidth}
    \includegraphics[width=\linewidth]{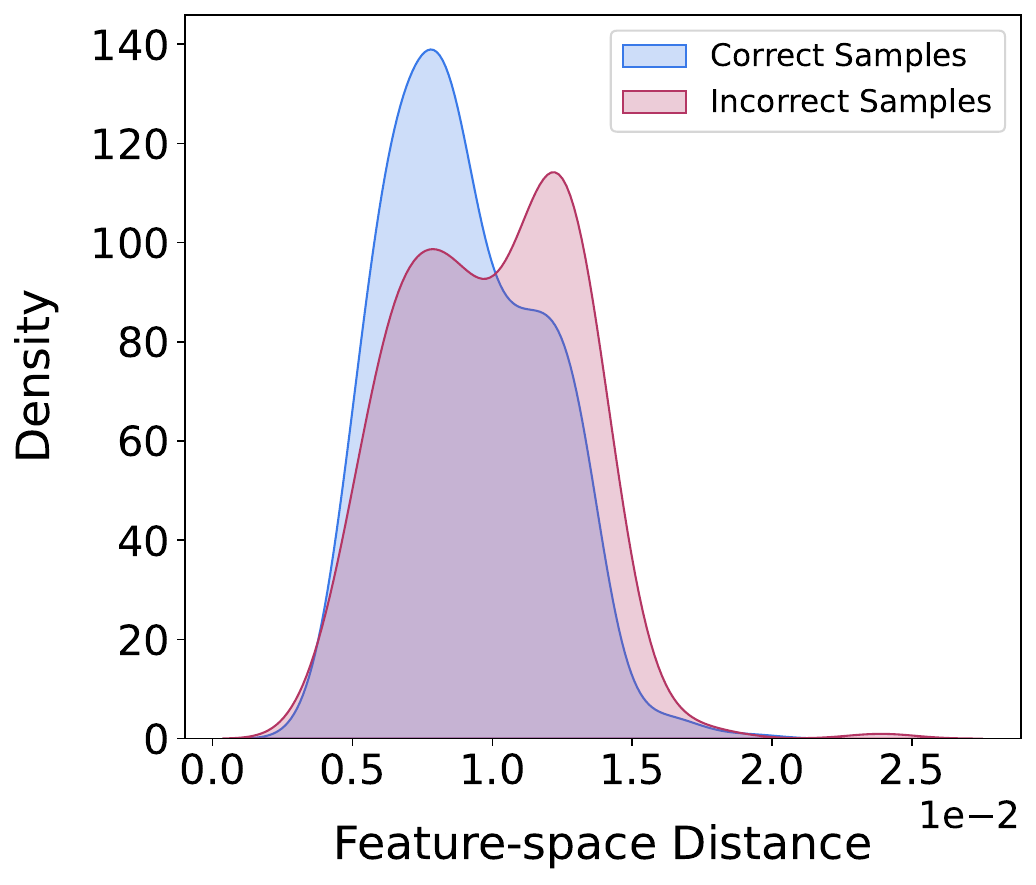}
    \caption{}       
    \label{fig:kde_incorrect}
    \end{subfigure}
    \begin{subfigure}[t]{0.35\linewidth}
    \includegraphics[width=\linewidth]{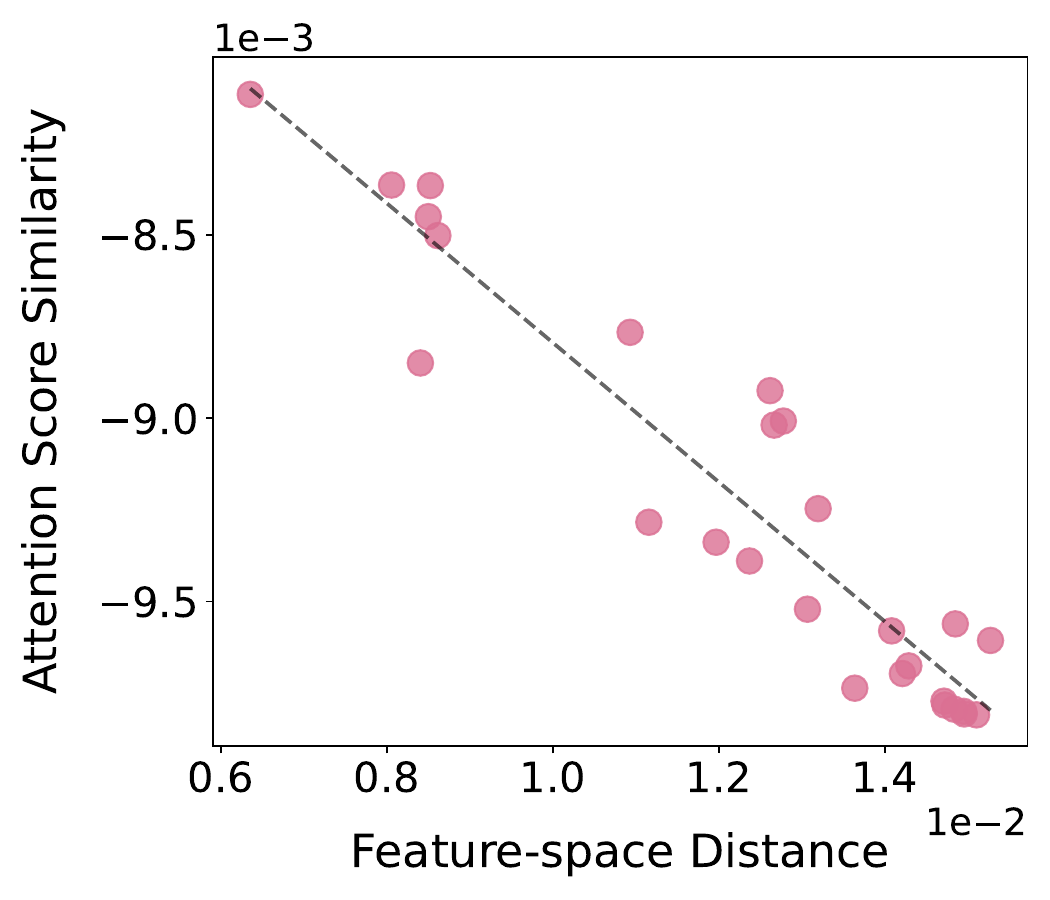}
    \caption{}
    \label{fig:attn_dist}
    \end{subfigure}
    \caption{\textbf{Motivating findings.} (a) Incorrect samples among Lens top-1 predictions form a bi-modal distribution. Larger mode is skewed toward the higher feature-space distances compared to the correct samples. (b) A strong correlation between attention score similarity and feature-space distance shows that samples closer in feature space exhibit more similar attention patterns.
    }
\end{figure}

\paragraph{Feature distance reveals failures of confidence-based selection.}
Confidence-based sensor control selects the physical capture with the highest model confidence~\cite{baek2025adaptive}.
However, high confidence does not necessarily imply that the selected capture is represented in a source-like feature space.
To examine this failure mode, we take the top-1 physical view selected by Lens on ImageNet-ES-Diverse~\cite{baek2025adaptive}, compute its feature-space distance to the pre-computed source-domain statistics, and split samples by whether the resulting prediction is correct.
As shown in Figure~\ref{fig:kde_incorrect}, incorrect predictions exhibit a bimodal distribution, with a larger mode shifted toward higher feature-space distances, whereas correct predictions are more concentrated at smaller distances.
This suggests that confidence-only selection can choose views that are confidently predicted but representation-shifted.

\paragraph{Feature distance tracks source-like attention.}
We further examine whether feature-space distance is related to the model's visual evidence.
For candidate physical views of the same scene, we compare their attention-score similarity to the corresponding source image.
Figure~\ref{fig:attn_dist} shows that views with smaller feature-space distance tend to have higher attention-score similarity.
Thus, source-affinity is not only a global feature-statistic criterion: it also correlates with whether the frozen VLM attends to visual evidence in a source-like manner.
This supports our design choice of extending the distribution-alignment principle of PromptAlign~\cite{abdul2023align} upstream, using source affinity to select physical measurements before post-capture aggregation.

\section{Physical Multiviews as an Augmentation Axis}
\label{appendix:physical_aug}

\begin{figure}
    \centering
    \includegraphics[width=0.6\linewidth]{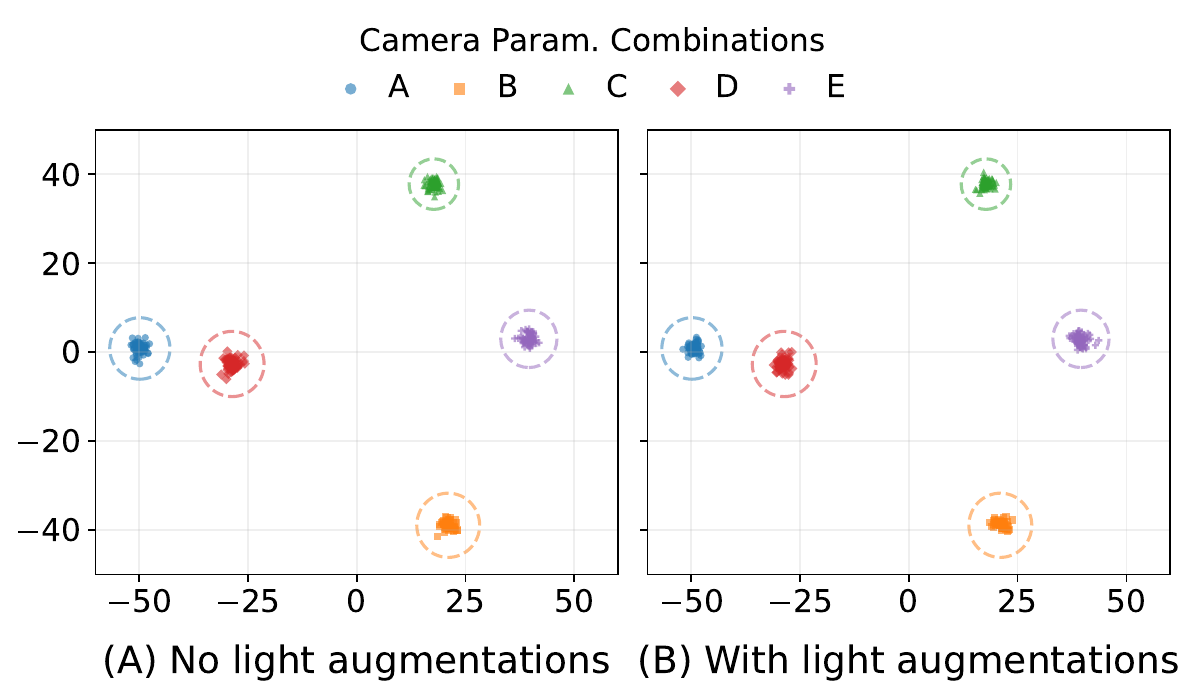}
    \caption{\textbf{LDA projection of visual embeddings from five physical views.}
    (Left) With digital transformation such as Random Crop or Horizontal Flip. (Right) With additional photometric augmentations such as perturbing hue, saturation, and brightness. Visualizations for more samples can be found in the Appendix.}
    \label{fig:orthogonal}
\end{figure}

\begin{figure*}[t]
    \centering
    \begin{subfigure}[b]{0.22\textwidth}
        \centering
        \includegraphics[width=\textwidth]{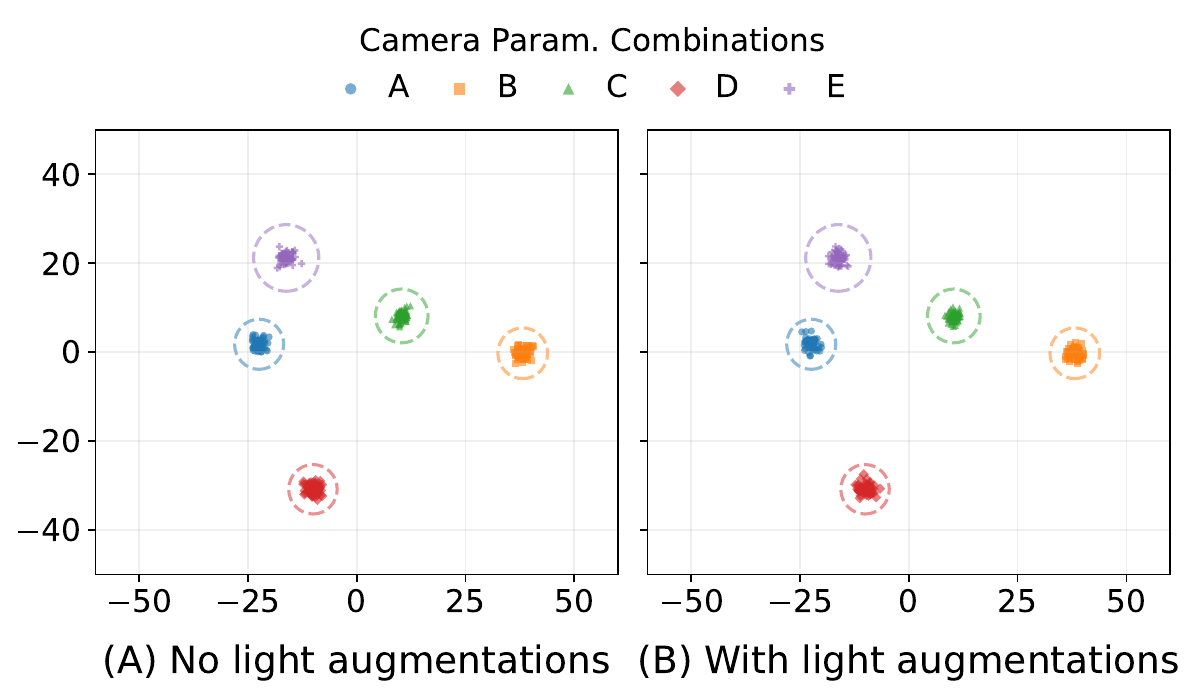}
        \caption{ImageNet-ES L1}
        \label{fig:appendix_LDA_ES_L1}
    \end{subfigure}
    \hfill
    \begin{subfigure}[b]{0.73\textwidth}
        \centering
        \vfill
        \includegraphics[width=\textwidth]{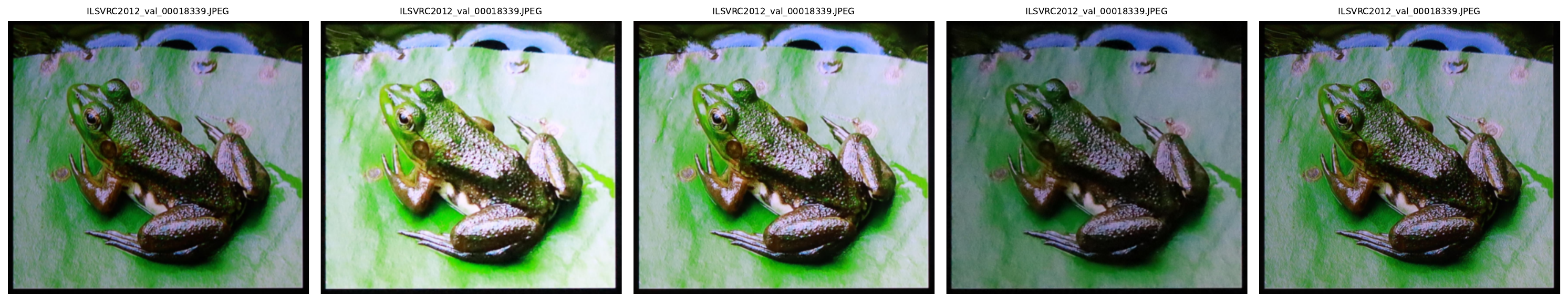}
        \caption{ImageNet-ES L1; American bullfrog}
        \label{fig:appendix_LDA_ES_L1_image}
    \end{subfigure}
    \hfill
    \begin{subfigure}[b]{0.22\textwidth}
        \centering
        \includegraphics[width=\textwidth]{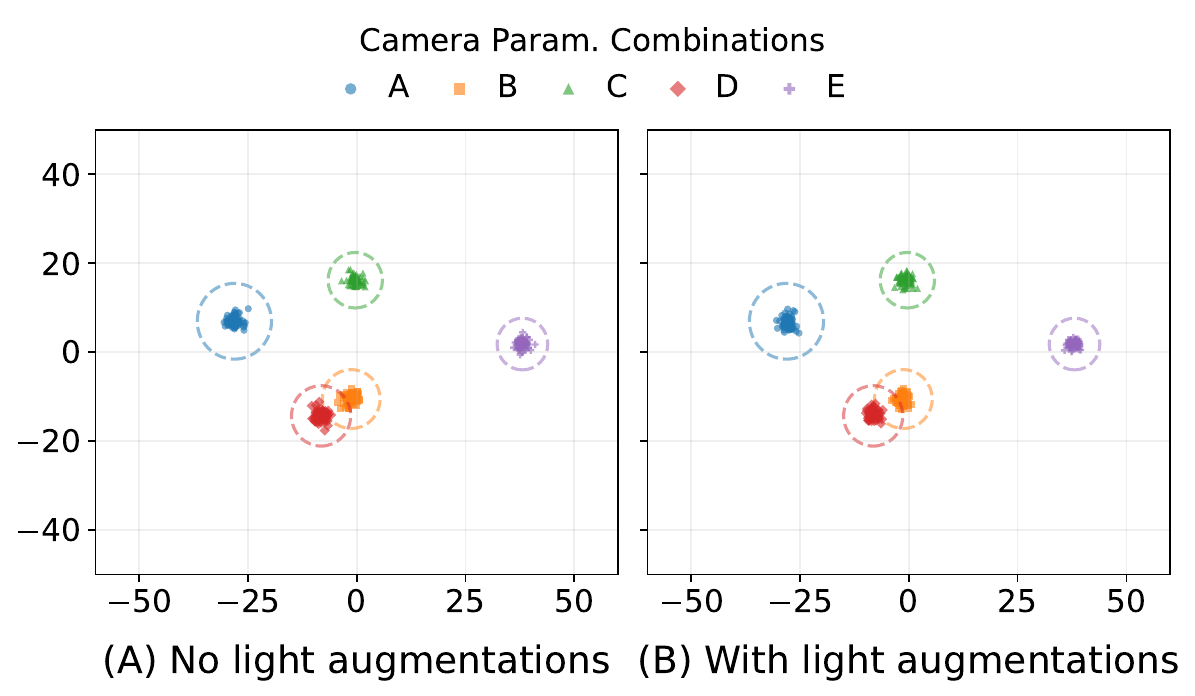}
        \caption{ImageNet-ES L5}
        \label{fig:appendix_LDA_ES_L5}
    \end{subfigure}
    \hfill
    \begin{subfigure}[b]{0.73\textwidth}
        \centering
        \vfill
        \includegraphics[width=\textwidth]{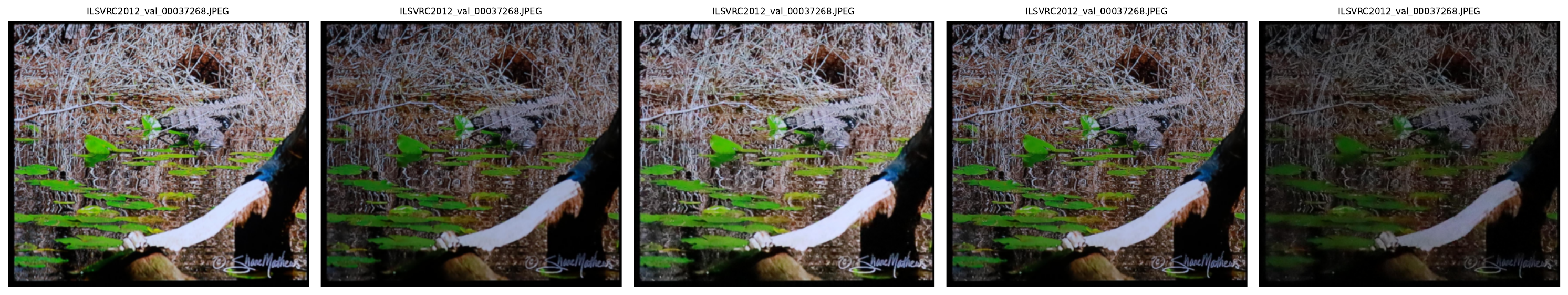}
        \caption{ImageNet-ES L5; American alligator}
        \label{fig:appendix_LDA_ES_L5_image}
    \end{subfigure}
    \hfill
    \begin{subfigure}[b]{0.22\textwidth}
        \centering
        \includegraphics[width=\textwidth]{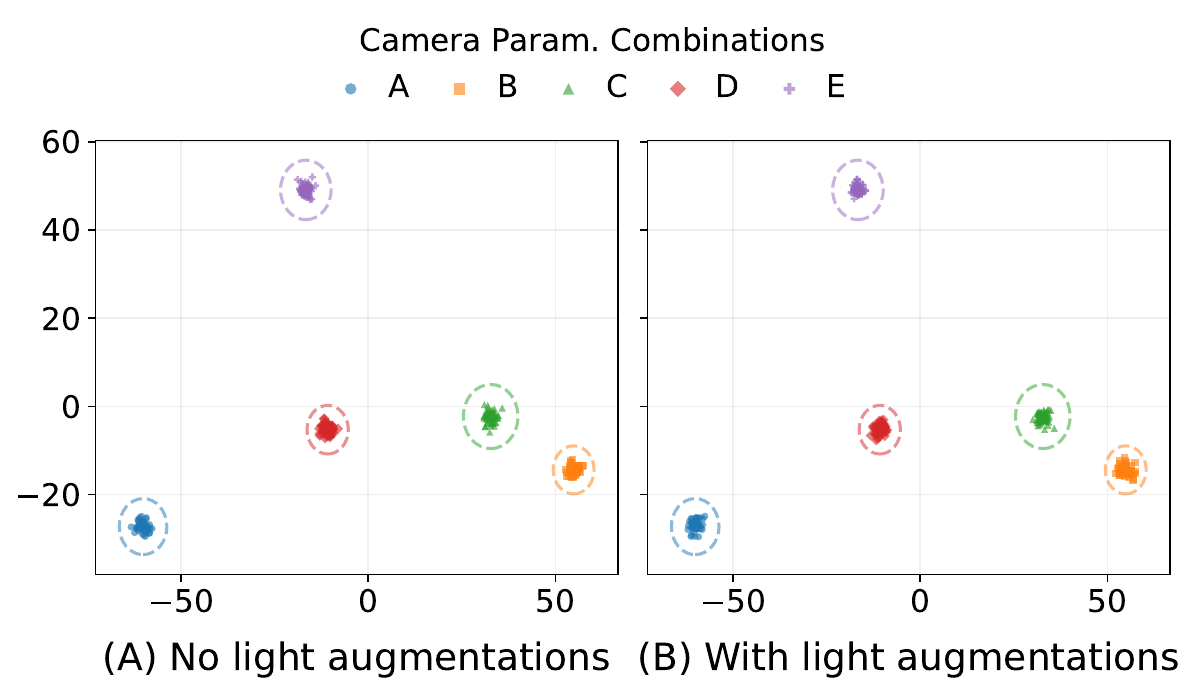}
        \caption{ImageNet-ES-Diverse L1}
        \label{fig:appendix_LDA_ES_Diverse_L1}
    \end{subfigure}
    \hfill
    \begin{subfigure}[b]{0.73\textwidth}
        \centering
        \vfill
        \includegraphics[width=\textwidth]{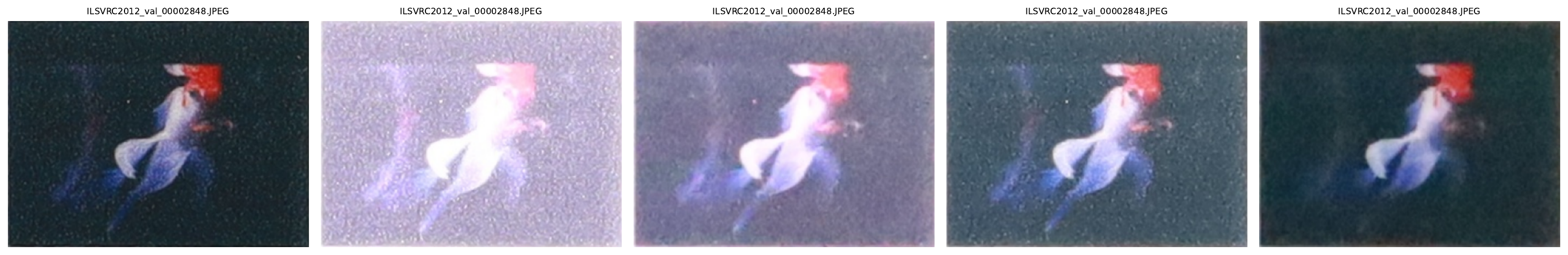}
        \caption{ImageNet-ES-Diverse L1; goldfish}
        \label{fig:appendix_LDA_ES_Diverse_L1_image}
    \end{subfigure}
    \hfill
    \begin{subfigure}[b]{0.22\textwidth}
        \centering
        \includegraphics[width=\textwidth]{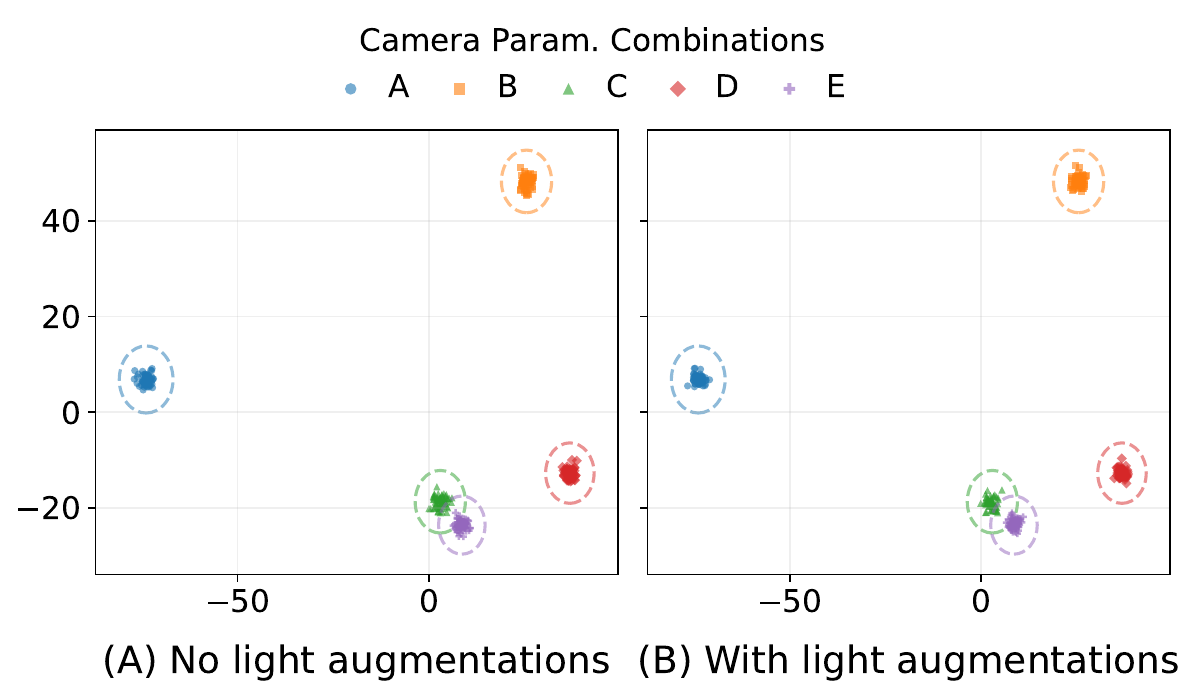}
        \caption{ImageNet-ES-Diverse L2}
        \label{fig:appendix_LDA_ES_Diverse_L2}
    \end{subfigure}
    \hfill
    \begin{subfigure}[b]{0.73\textwidth}
        \centering
        \vfill
            \includegraphics[width=\textwidth]{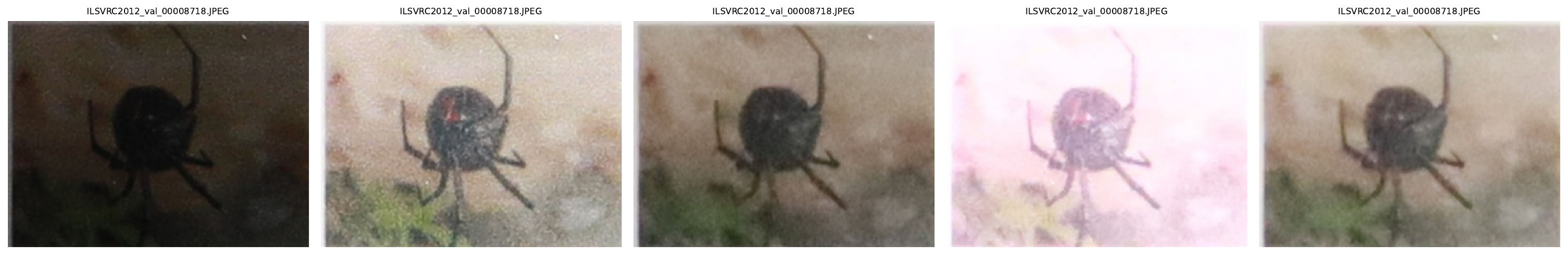}
        \caption{ImageNet-ES-Diverse L2; southern black widow}
        \label{fig:appendix_LDA_ES_Diverse_L2_image}
    \end{subfigure}
    \hfill
    \begin{subfigure}[b]{0.22\textwidth}
        \centering
        \includegraphics[width=\textwidth]{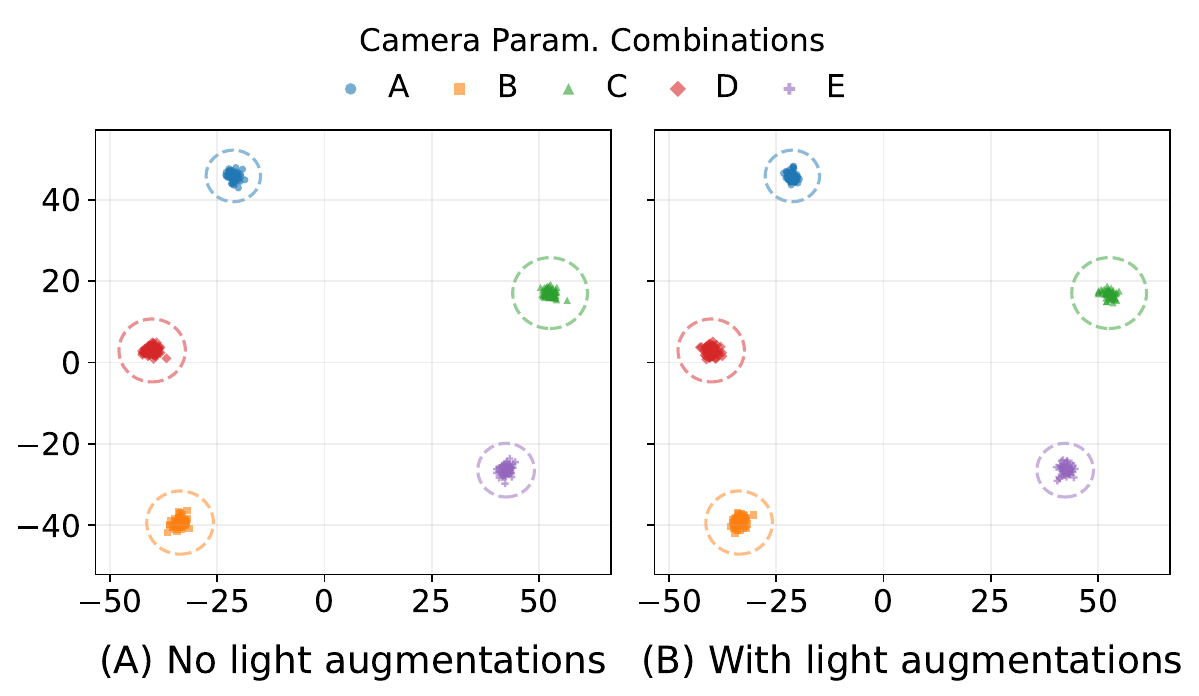}
        \caption{ImageNet-ES-Diverse L3}
        \label{fig:appendix_LDA_ES_Diverse_L3}
    \end{subfigure}
    \hfill
    \begin{subfigure}[b]{0.73\textwidth}
        \centering
        \vfill
        \includegraphics[width=\textwidth]{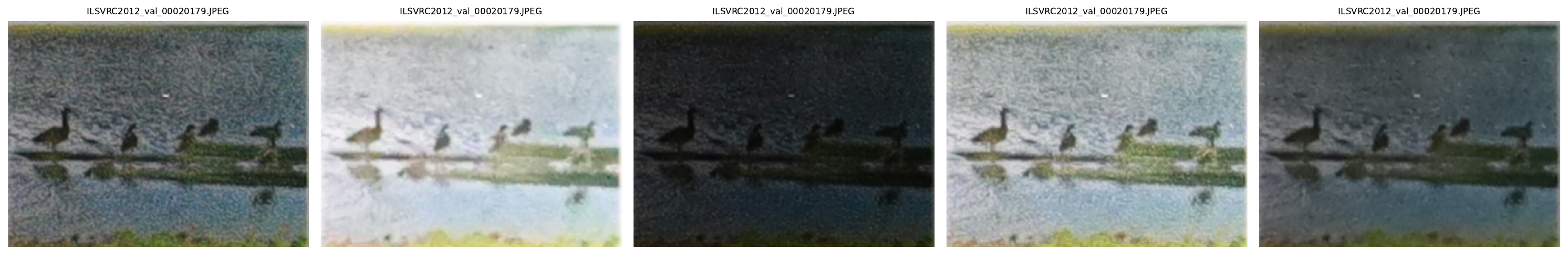}
        \caption{ImageNet-ES-Diverse L3; goose}
        \label{fig:appendix_LDA_ES_Diverse_L3_image}
    \end{subfigure}
    \hfill
    \begin{subfigure}[b]{0.22\textwidth}
        \centering
        \includegraphics[width=\textwidth]{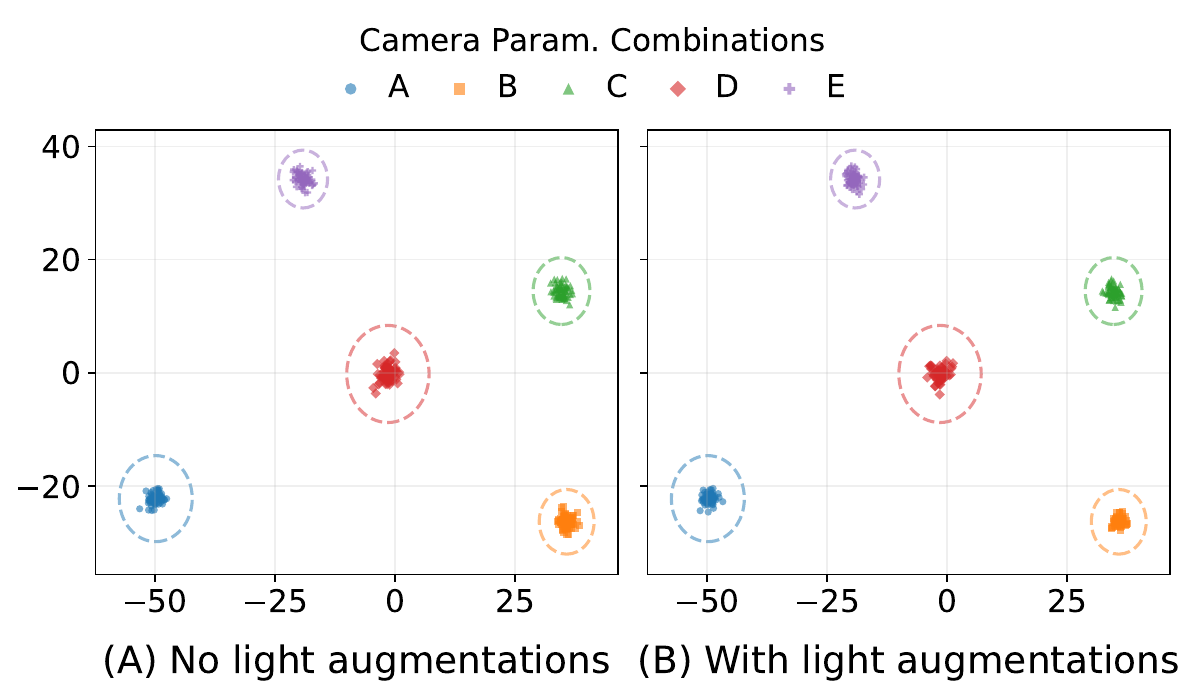}
        \caption{ImageNet-ES-Diverse L4}
        \label{fig:appendix_LDA_ES_Diverse_L4}
    \end{subfigure}
    \hfill
    \begin{subfigure}[b]{0.73\textwidth}
        \centering
        \vfill
        \includegraphics[width=\textwidth]{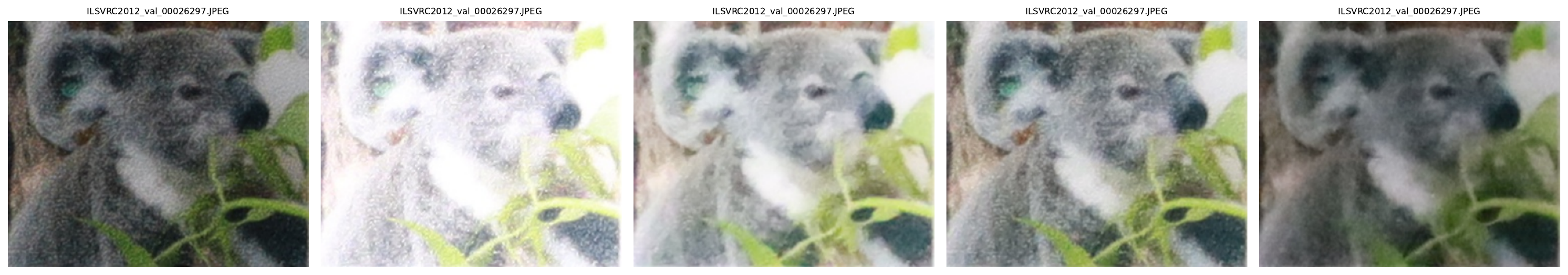}
        \caption{ImageNet-ES-Diverse L4; koala}
        \label{fig:appendix_LDA_ES_Diverse_L4_image}
    \end{subfigure}
    \hfill
    \begin{subfigure}[b]{0.22\textwidth}
        \centering
        \includegraphics[width=\textwidth]{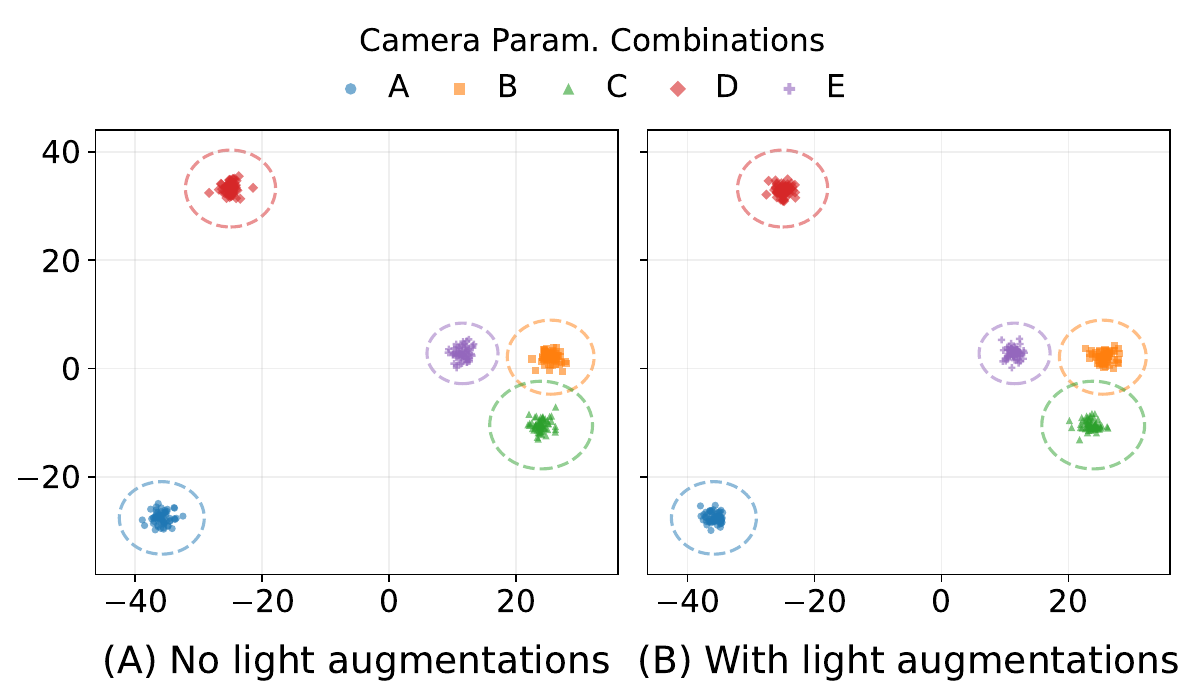}
        \caption{ImageNet-ES-Diverse L6}
        \label{fig:appendix_LDA_ES_Diverse_L6}
    \end{subfigure}
    \hfill
    \begin{subfigure}[b]{0.7\textwidth}
        \centering
        \vfill
        \includegraphics[width=\textwidth]{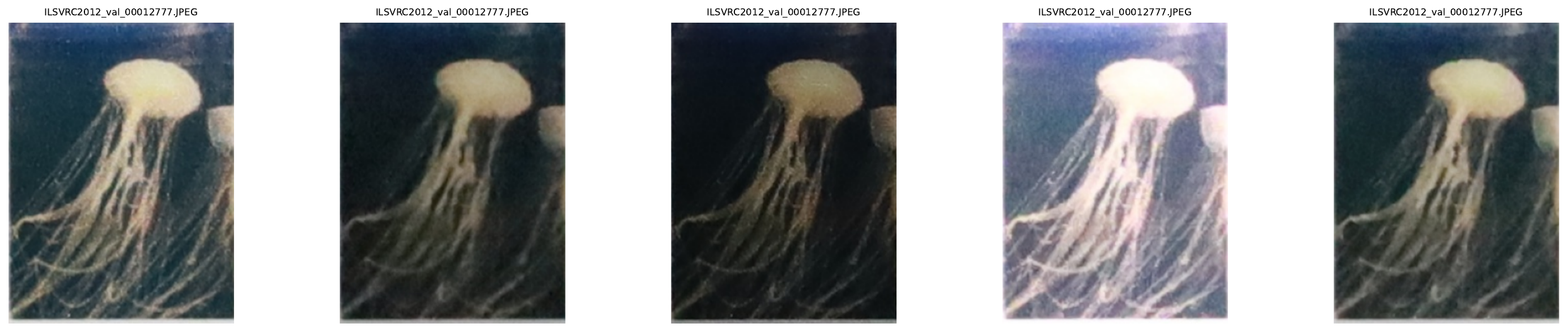}
        \caption{ImageNet-ES-Diverse L6; jellyfish}
        \label{fig:appendix_LDA_ES_Diverse_L6_image}
    \end{subfigure}
    \hfill
    \begin{subfigure}[b]{0.22\textwidth}
        \centering
        \includegraphics[width=\textwidth]{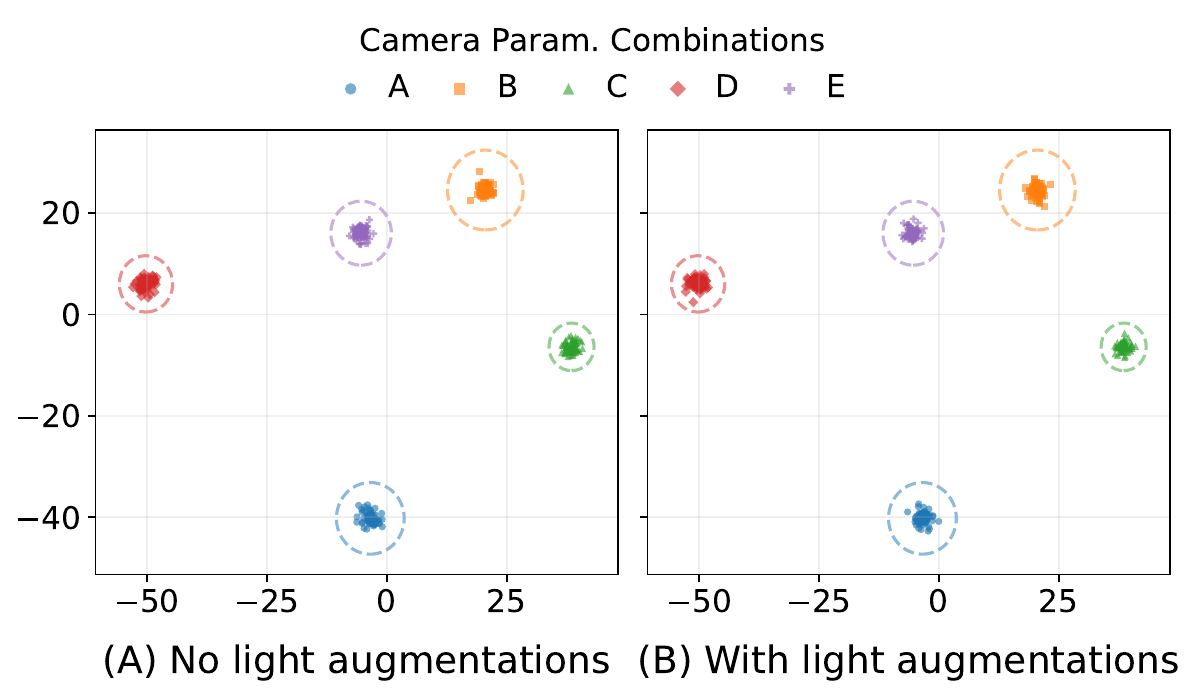}
        \caption{ImageNet-ES-Diverse L7}
        \label{fig:appendix_LDA_ES_Diverse_L7}
    \end{subfigure}
    \hfill
    \begin{subfigure}[b]{0.73\textwidth}
        \centering
        \vfill
        \includegraphics[width=\textwidth]{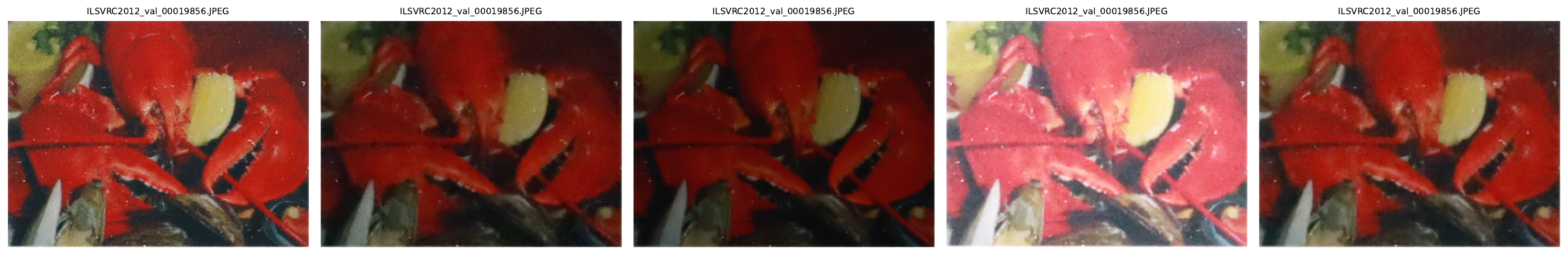}
        \caption{ImageNet-ES-Diverse L7; American lobster}
        \label{fig:appendix_LDA_ES_Diverse_L7_image}
    \end{subfigure}
    \vspace{-1ex}
    \caption{Additional LDA Projection Visualizations by Light Condition}
    \label{fig:appendix_additional_LDAs}
\end{figure*}

We further examine whether physical multiviews provide feature-space variability that is distinct from standard digital augmentations.
For each scene, we take the top-5 physical views selected by source affinity and project their digitally augmented visual embeddings using LDA.
As shown in Figure~\ref{fig:orthogonal}, embeddings from different sensor settings occupy distinct regions, whereas geometric augmentations remain largely within each sensor-specific region.
Adding photometric perturbations~\cite{buslaev2020albumentations} changes appearance but still does not fully reproduce the feature-space variability induced by changing camera parameters.
Figure~\ref{fig:appendix_additional_LDAs} shows additional LDA projection visualizations for each lighting condition in ImageNet-ES and ImageNet-ES-Diverse, along with the corresponding top-$5$ camera parameters ranked by affinity score.
This supports our use of multiple physical captures as a complementary augmentation axis for hard-vote aggregation.

\section{ImageNet-ES/ImageNet-ES-Diverse details}
\label{appendix:dataset}
ImageNet-ES~\cite{baek2024unexplored} and ImageNet-ES-Diverse~\cite{baek2025adaptive} are benchmarks designed to evaluate robustness under sensor-level covariate shifts induced by variations in lighting conditions and camera parameters. Their test sets consist of 64k and 192k samples, respectively.

The two datasets differ primarily in object presentation: ImageNet-ES captures a monitor displaying the objects, treating them as light-emitting sources, while ImageNet-ES-Diverse captures printed images of the objects, enabling more realistic evaluation under ambient lighting. Furthermore, ImageNet-ES includes only two lighting conditions (e.g. light-on, light-off), whereas ImageNet-ES-Diverse provides six diverse illumination settings. 

\begin{figure*}
    \centering
    \begin{subfigure}[b]{0.33\textwidth}
        \centering
        \includegraphics[width=\textwidth]{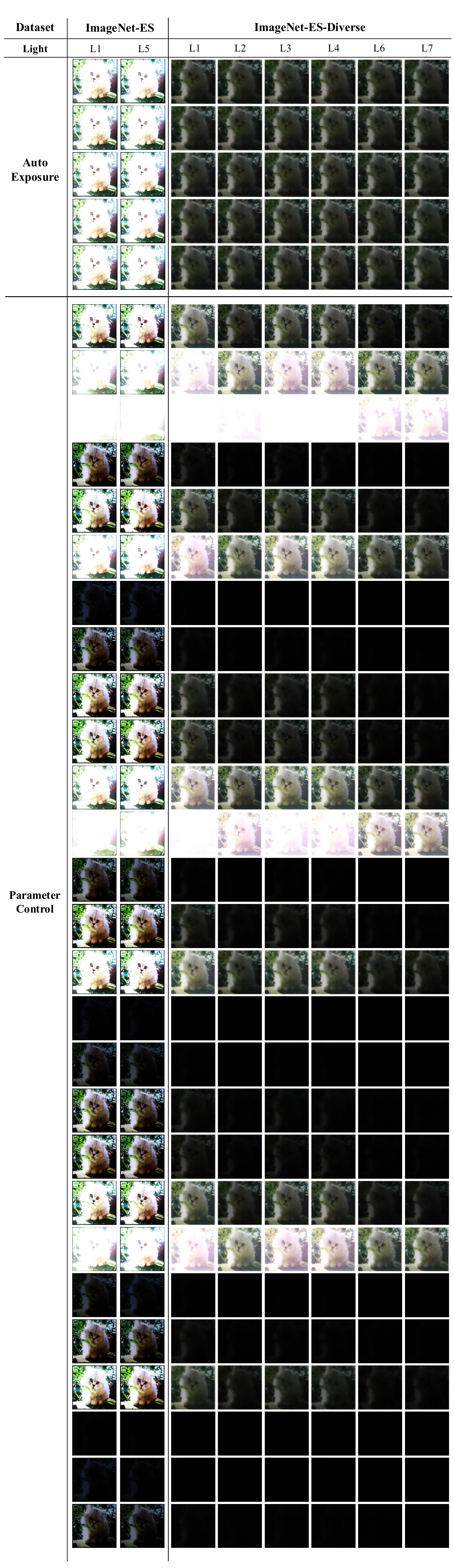}
        \captionsetup{margin={2em,0em}}
        \caption{Class name: Persian cat}
        \label{fig:appendix_es_1}
    \end{subfigure}
    \begin{subfigure}[b]{0.33\textwidth}
        \centering
        \includegraphics[width=\textwidth]{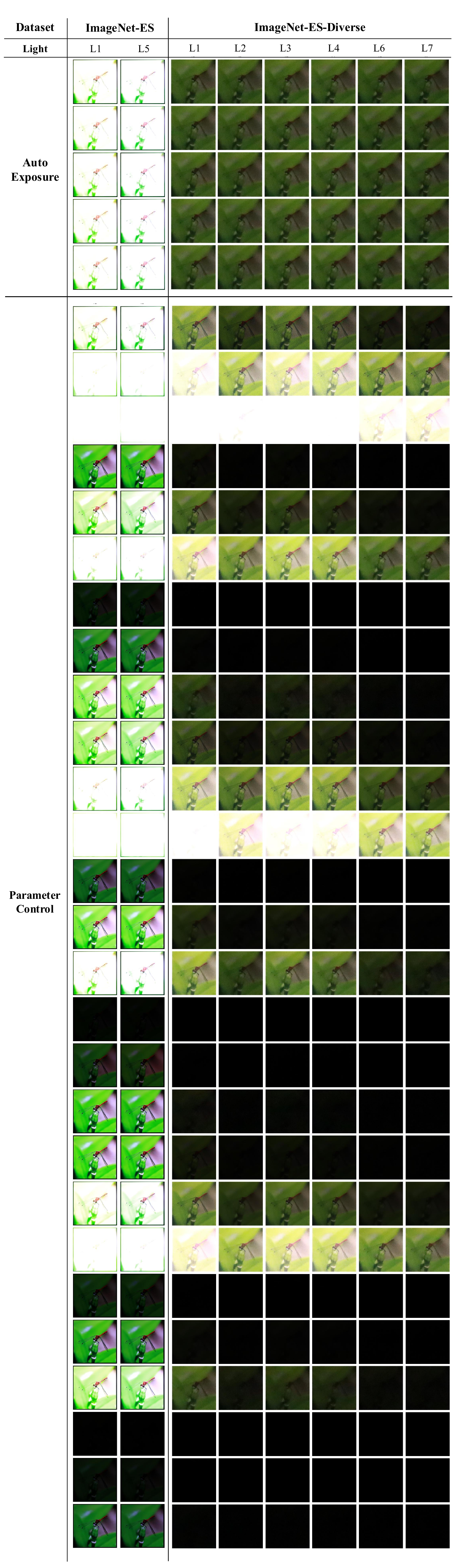}
        \captionsetup{margin={2em,0em}}
        \caption{Class name: Dragonfly}
        \label{fig:appendix_es_2}
    \end{subfigure}
    \begin{subfigure}[b]{0.33\textwidth}
        \centering
        \includegraphics[width=\textwidth]{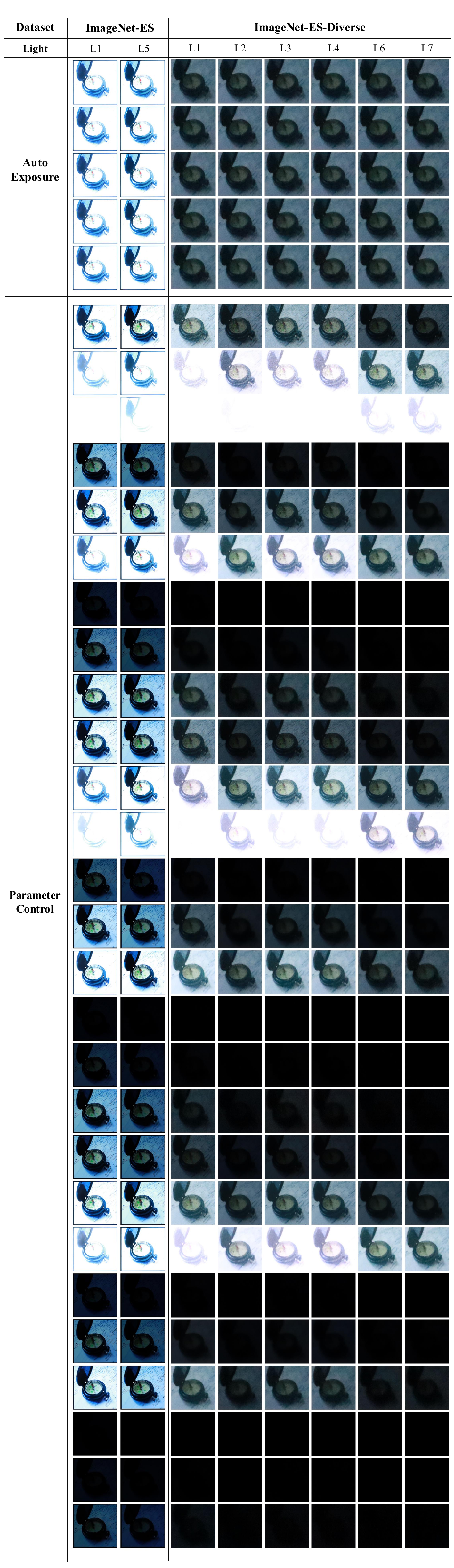}
        \captionsetup{margin={2em,0em}}
        \caption{Class name: Magnetic compass}
        \label{fig:appendix_es_3}
    \end{subfigure}
    \caption{Additional qualitative results}
    \label{fig:dataset_samples}
\end{figure*}

Each dataset contains (i) an \textit{Auto-Exposure (AE)} subset, where scenes are captured several times using the camera’s default exposure setting, and 
(ii) a \textit{Parameter-Controlled} subset, where the same scenes are recorded under systematically varied ISO, shutter speed, and aperture combinations across multiple lighting environments.
The AE set contains five captures per scene using the camera’s automatic exposure mode, while the parameter-controlled set contains 27 captures obtained under controlled combinations of ISO, shutter speed, and aperture—three levels each (ISO: \{250, 2000, 16000\}, Shutter: \{1/4$^{\prime\prime}$, 1/60$^{\prime\prime}$, 1/1000$^{\prime\prime}$\}, Aperture: {f/5.0, f/9.0, f/16}).
Example samples from both datasets are shown in Fig.~\ref{fig:dataset_samples}.

\section{Implementation details and full results about main evaluation} 
\label{appendix:implementation}
\paragraph{Details of \ours}
We implement \ours~upon MaPLe~\cite{khattak2023maple}, the multi-modal prompting model with the CLIP ViT-B/16 backbone. 
Source-affinity score is computed for a depth of three layers of the visual encoder $(L=3)$, while the source dataset statistics are pre-computed on ImageNet. 
For each physical view, we generate 63 digitally augmented views using random resized cropping and horizontal flipping, forming a batch of $N=64$ digitally augmented images including the original as in prior TTA methods. 
We use the fraction $\alpha=0.3$ to include only confident augmentation samples in source-affinity score aggregation. 
We select the top-5 parameter settings (i.e., $k$=5) with the largest source-affinity score, and within their augmented samples, retain the bottom $\gamma$=3\% in entropy as confident predictions for voting. 

\paragraph{Details of baselines}
We compare \ours~to representative approaches across three categories. (1) Prompt learning methods, including CoOp~\cite{zhou2022learning} and MaPLe~\cite{khattak2023maple}, which optimize learnable context tokens for downstream adaptation. We use the publicly released pretrained weights of CoOp and MaPLe without any additional training. (2) Prompt-based TTA methods, including TPT~\cite{shu2022test}, C-TPT~\cite{sharifdeen2025tpt}, O-TPT~\cite{yoon2024c}, and PromptAlign~\cite{abdul2023align}, which adapt prompt representations at test time. (3) Training-free, forward-only TTA methods, including MTA~\cite{zanella2024test}, TDA~\cite{karmanov2024efficient}, and ZERO~\cite{farina2024frustratingly}. In addition, we include Lens~\cite{baek2025adaptive} as a sensor-control baseline, which selects the single most confident physical observation for each scene based on model confidence. A more detailed discussion of Lens and its comparison with \ours~is provided in Section~\ref{appendix:lens}.

To ensure a fair comparison, all TTA baselines are initialized with the same pretrained MaPLe~\cite{khattak2023maple} weights, which place learnable prompt vectors in both the text and visual encoders up to the third layer. Following each method's original design, we restrict gradient updates to the textual prompt vectors (\texttt{prompt\_learner.ctx}) for all prompt-based TTA baselines except PromptAlign. PromptAlign is the only method allowed to update visual prompt vectors, as its objective explicitly relies on adapting visual features for cross-modal alignment. Consequently, for all other prompt-based TTA methods, no gradients flow into the visual encoder and only the text-encoder input prompts are optimized during test-time adaptation.

\paragraph{Full results}
We also report the full accuracy results across individual lighting environments for both ImageNet-ES~\cite{baek2024unexplored} and ImageNet-ES-Diverse~\cite{baek2025adaptive}.
Table~\ref{tab:main_detail} summarizes detailed accuracies under the three evaluation settings used in the main paper—AE, AE + photometric augmentations, and Lens-based selection—across all lighting conditions.

\section{Effect of Proxy Dataset Choice}
\label{appendix:proxy}
Since both ImageNet-ES and ImageNet-ES-Diverse are derived from ImageNet, we additionally use LAION~\cite{schuhmann2021laion}, a large-scale and diverse web-scale dataset, as the proxy dataset for computing source-affinity scores. To isolate the effect of the proxy dataset, all experiments are conducted using the same CLIP ViT-B/16 backbone without MaPLe initialization. As shown in Table~\ref{tab:appendix_proxy}, replacing ImageNet with LAION results in a slight performance decrease. Nevertheless, the performance drop only remains within approximately 1.5\%p on ImageNet-ES and 0.5\%p on ImageNet-ES-Diverse, indicating that the proposed source-affinity mechanism still remains effective under an alternative proxy distribution.

\begin{table*}[t]
    \centering
    \caption{\textbf{Evaluation with Alternative Proxy Datasets.} Source-affinity scores are computed using either ImageNet or LAION as the proxy dataset. To ensure a fair comparison, all experiments use the same CLIP ViT-B/16 backbone without MaPLe prompts.}
    \label{tab:appendix_proxy}
    \begin{tabular}{c|ccc|ccccccc}
        \toprule
       \multirow{2}{*}{\textbf{\makecell{Proxy\\Dataset}}} & \multicolumn{3}{c|}{{ImageNet-ES}} & \multicolumn{7}{c}{{ImageNet-ES-Diverse}} \\
        \cmidrule(lr){2-4} \cmidrule(lr){5-11}
        & {L1} & {L5} & {Average} & {L1} & {L2} & {L3} & {L4} & {L6} & {L7} & {Average} \\
        \midrule
        {ImageNet} & 85.3 & 84.3 & 84.80 & 65.3 & 64.8 & 65.5 & 64.2 & 63.4 & 62.7 & 64.32 \\
        {LAION} & 82.4 & 82.5 & 82.45 & 63.2 & 65.3 & 64.2 & 63.4 & 64.0 & 63.0 & 63.85 \\
    \bottomrule
    \end{tabular}
\end{table*}

\section{Conceptual Difference from Lens}
\label{appendix:lens}
Lens and MVP share a common hardware-driven philosophy: rather than modifying the model itself through additional training or adaptation, both approaches aim to improve performance by providing the model with more informative physical observations.

However, the two approaches operate at different levels. Lens performs confidence-based selection of a single physical observation and directly uses the selected image for prediction. In contrast, MVP employs source-affinity-based selection as the first stage of a broader multi-view adaptation framework. Rather than making a prediction from a single selected observation, MVP first identifies multiple source-aligned physical observations and subsequently aggregates them through entropy-based filtering and voting.

Consequently, Lens can be viewed as a view-selection strategy that is compatible with existing test-time adaptation methods, similar to the setting explored in our experiments, as mentioned in~\ref{appendix:implementation}. In MVP, by contrast, view selection constitutes only one component within the overall adaptation pipeline, which further incorporates adaptation and multi-view aggregation to produce the final prediction.

\section{Evaluation on different architectures}
\label{appendix:arch}
To verify that the effectiveness of \ours~does not depend on a specific prompting architecture, we further evaluate our framework on two additional backbones: 
(1) CLIP ViT-B/16 \emph{without} MaPLe initialization (\textit{i.e.,} no visual prompt tokens), and 
(2) CLIP ResNet50. 
All components of \ours~—including feature-space distance computation, physical-view selection, and entropy-filtered aggregation—are kept identical to the main setup; only the underlying visual encoder is replaced. For fair comparison, the text prompt is initialized using the standard CLIP template, “a photo of a {class}”, without any learned prompt vectors.

For ViT-B/16 \textit{without} MaPLe, we retain the identical procedure to the main experiments for computing source statistics and feature-space distances.

For ResNet50, we replace the layer-wise token statistics used for ViT with block-wise statistics by treating each \texttt{Bottleneck block} as an individual layer.
Specifically, each block’s feature map $[B,C,H,W]$ is reshaped by flattening the spatial dimensions $(H\times W)$ so that each spatial location is regarded as a token.
We then compute mean and variance over the resulting (token, channel) representations and measure the feature-space affinity  between these block-level statistics and the pre-computed source statistics. The affinity score is computed over the first 7 out of 16 bottleneck blocks, following the observation in~\cite{raghu2021vision} that ResNet requires a larger portion of its early blocks to produce representations comparable to those obtained from a much smaller number of early ViT layers.

Results are presented in Table~\ref{tab:vit_no_maple} and~\ref{tab:rn50}.
Across both backbones, the average performance of \ours~consistently outperforms all forward-only and prompt-based TTA baselines, demonstrating that its effectiveness does not depend on the choice of visual backbone or on the presence of tuned prompt vectors.

\section{Qualitative analyses.}
\label{appendix:qual}
\begin{figure*}
    \centering
    \includegraphics[width=0.9\linewidth]{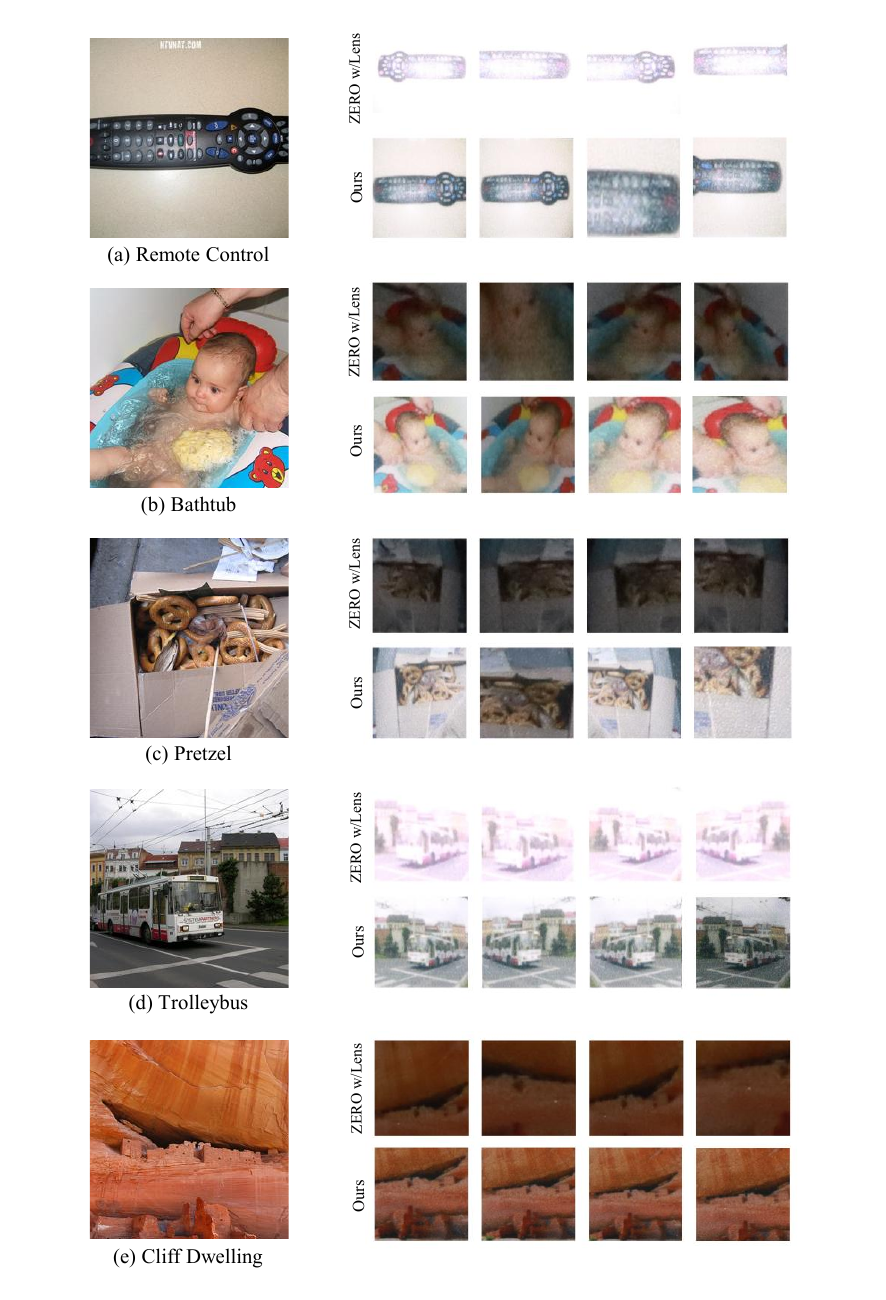}
    \vspace{-4.5ex}
    \caption{Additional qualitative results}
    \label{fig:appendix_qual}
    \vspace{-3ex}
\end{figure*}

Figure~\ref{fig:appendix_qual} presents confident, digitally-augmented views selected for the final prediction by ZERO with Lens and \ours, respectively.
\ours~avoids being trapped by mistakenly selected overconfident physical view 
in the capture time by considering various physical settings in terms of the model's perception and delivers optically improved images. It is noteworthy that the lack of the diversity introduced by a limited physical sensor space in a conventional sensor-control approach cannot be mitigated solely by digitally augmented views. 


\section{Ablation Study}
\label{appendix:ablation}

\begin{figure}[h]
    \centering
    \begin{subfigure}{0.42\linewidth}
        \includegraphics[width=\linewidth]{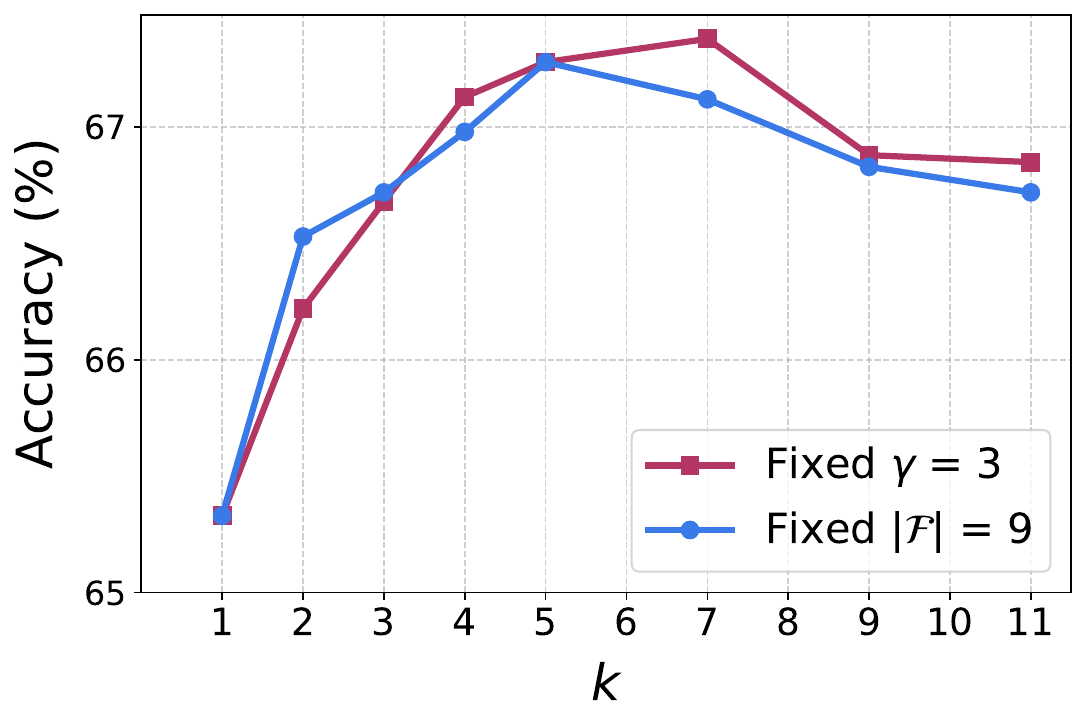}
        \caption{Effect of various $k$}
        \label{fig:topk}
    \end{subfigure}
    \begin{subfigure}{0.42\linewidth}
        \includegraphics[width=\linewidth]{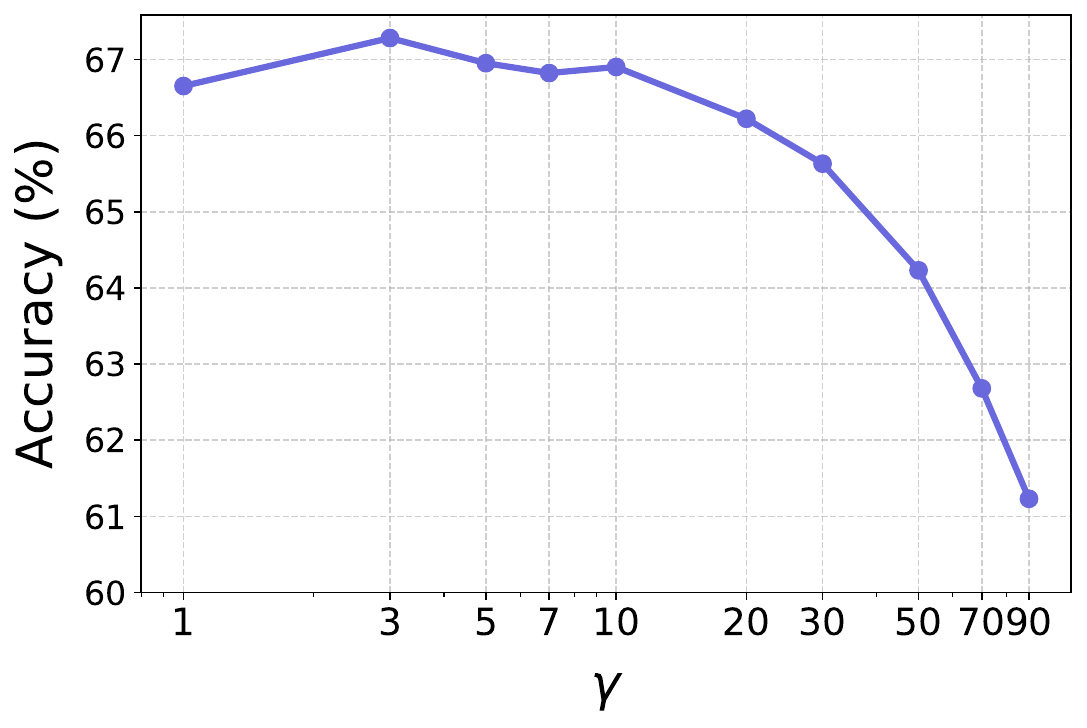}
        \caption{Effect of various $\gamma$}
        \label{fig:percentile}
    \end{subfigure}
    \caption{\textbf{Ablation on parameter-view selection in \ours.} (a) shows the selection of top-$k$ physical parameter settings based on the source-affinity score, while (b) illustrates the confidence percentile thresholding used to filter augmented views for final prediction. We fix $k=5$ and vary $\gamma$ from 1\% to 90\%.}
    \label{fig:ablation_1}
\end{figure}

\paragraph{Effect of varying $k$.}
We examine how the number of physical views selected by the source‑affinity score affects accuracy.
Figure~\ref{fig:topk} shows an inverted U‑shaped relationship between $k$ and performance. With a fixed confidence percentile $\gamma=3\%$ (red), increasing $k$ raises the total number of augmented views that enter the vote; when $k$ is too small, the ensemble has too few voters, leading to unstable predictions.
To decouple the effect of $k$ from the number of voters, we also hold the total count of selected augmented views constant at $|\mathcal{F}|=9$ (blue). 
In this setting, smaller $k$ allocates more augmentations per physical view, allowing lower‑confidence augmentations to participate in the vote. Conversely, once $k$ exceeds the plateau, accuracy declines slightly because additional, lower–source‑affinity physical views are admitted and their associated (often overconfident) augmentations introduce noise that dilutes the vote.



\paragraph{Effect of varying $\gamma$.} Figure~\ref{fig:percentile} reports the effect of varying the entropy filtering threshold $\gamma$, which determines the proportion of digitally augmented views selected for the final aggregation. 
As shown, the performance peaks at $\gamma$ = 3\% and remains relatively stable up to $\gamma$ = 10\%, after which it sharply degrades, indicating that including too many high-entropy (less confident) views introduces noise into the aggregation process, resulting in unstable predictions.

\begin{table}[h]
    \centering
    \caption{\textbf{Ablation Study on Physical Augmentation Design Choices.} Comparison across digital (geometric/photometric) augmentations and physical augmentations using multiple camera parameter views. $k$ physical views are selected from all 27 camera parameter settings in each case. A total of 320 augmented views are produced per scene for $k{=}1$ settings, matching the number of those at $k{=}5$ setting; $64 \times 5 = 320$ views. Therefore, all comparisons are conducted using the same pool of pre-captured physical observations, with differences arising only from the view selection and aggregation strategy.}
    \label{tab:key_ablation}
    \setlength{\tabcolsep}{1mm} 
    \fontsize{9pt}{11pt}\selectfont
        \begin{tabular}{cccc|cc}
            \toprule
            \textbf{\makecell{Number of top  \\physical views ($k$)}} & \textbf{\makecell{Digital\\geo. aug.}} & \textbf{\makecell{Digital\\photo. aug.}} & \textbf{\makecell{Physical\\aug.}}  & \textbf{IN-ES} & \textbf{\makecell{IN-ES\\-Diverse}} \\
            \midrule
            \makecell{$k=1$} & \makecell{\checkmark} & \makecell{} & \makecell{} & 86.90 & 66.72 \\
            \makecell{$k=1$} & \makecell{\checkmark} & \makecell{\checkmark} & \makecell{} & 86.85 & 66.62 \\
            \midrule
            \makecell{Ours ($k=5$)} & \makecell{\checkmark} & \makecell{} & \makecell{\checkmark} & \textbf{87.85} & \textbf{67.28} \\
            \bottomrule
        \end{tabular}
\end{table}

\paragraph{Number of physical views.} 
Table~\ref{tab:key_ablation} highlights the importance of leveraging multiple physical views rather than relying on a single capture or digital-level augmentations. 
Aggregating predictions from multiple physical captures consistently outperforms applying digital geometric or additional photometric augmentations to a single physical capture. 
This demonstrates that using multiple physical views leads to more robust and stable predictions than depending on a single parameter setting. 
Moreover, even when photometric augmentations are added to the single physical view, the performance still falls short of our approach, underscoring the inherent limitation of digital-level augmentations in simulating sensor-level diversity.

\paragraph{Component-wise Ablation Study.} Table~\ref{tab:appendix_keys} presents an ablation study on the key components of our framework. We first evaluate different physical view selection strategies. Compared with random selection, source-affinity-based selection consistently achieves higher performance on both benchmarks. This result highlights the importance of sensor control in physical multi-view adaptation. Rather than relying on arbitrary camera parameter configurations, selecting physical views that are better aligned with the source distribution provides observations that are more suitable for the model, leading to more reliable adaptation.

We further compare source-affinity-based selection against using all available physical views without sensor control. Although leveraging all views already provides strong performance, source-affinity-based selection consistently achieves better results. This indicates that not all camera parameter configurations contribute equally to adaptation and that filtering physical views according to their source affinity produces a more reliable candidate set for downstream prediction.

Finally, we compare two aggregation strategies using the selected and entropy-filtered views. While marginalized aggregation averages prediction probabilities across views, hard voting determines the final prediction based on label-level consensus. As shown in Table~\ref{tab:appendix_keys}, hard voting consistently outperforms marginalized aggregation, suggesting that consensus-based aggregation is more robust to noisy or miscalibrated predictions from individual views.

Combining source-affinity-based physical view selection, entropy-based digital view filtering, and hard voting achieves the best overall performance on both benchmarks, validating the effectiveness of each component in the proposed framework.

\begin{table}[h]
    \centering
    \caption{\textbf{Ablation Study on Key Components.} Performance impact of physical view selection and prediction aggregation strategies. "No Sensor Control" applies digital view filtering and hard voting over all physical views. "Random Selection" randomly selects $k$ physical views before digital filtering and hard voting, and the reported results are averaged over three random seeds. "Marginalization" replaces hard voting with probability averaging over source-affinity-selected and entropy-filtered views. The proposed framework combines source-affinity-based physical view selection, digital view filtering, and hard voting, achieving the best performance on both benchmarks.}
    \label{tab:appendix_keys}
    \setlength{\tabcolsep}{1mm} 
    \fontsize{9pt}{11pt}\selectfont
        \begin{tabular}{cccc|cc}
            \toprule
            \textbf{\makecell{Method}} & \textbf{\makecell{Physical View \\ Selection}} & \textbf{\makecell{Digital View \\ Filtering}} & \textbf{\makecell{Hard \\ Voting}}  & \textbf{IN-ES} & \textbf{\makecell{IN-ES\\-Diverse}} \\
            \midrule
            \makecell{No Sensor Control} & \makecell{All views} & \makecell{\checkmark} & \makecell{\checkmark} & 87.25 & 66.37 \\
            \makecell{Random Selection} & \makecell{Random $k$ views} &
            \makecell{\checkmark} & \makecell{\checkmark} & 86.52 & 61.90
            \\
            \makecell{Marginalization} & \makecell{Source Affinity} & \makecell{\checkmark} & \makecell{} & 87.80 & 66.72 \\
            \midrule
            \makecell{\textbf{Ours}} & \makecell{Source Affinity} & \makecell{\checkmark} & \makecell{\checkmark} & \textbf{87.85} & \textbf{67.28} \\
            \bottomrule
        \end{tabular}
\end{table}

\paragraph{Source-affinity score measurement layers.} We evaluate the overall performance by varying the choice of layers for measuring feature-space source affinity score, as summarized in Table~\ref{tab:layer_ablation}.
It shows that using early layers, particularly the first to third layers, achieves the best accuracy on both datasets. 
In contrast, computing the score over all layers or only the later layers leads to a significant performance degradation. 
By measuring the affinity in early-layer feature space, our method can more reliably identify physical views that the model perceives as closer to its pre-training domain—effectively providing inputs that are “familiar” to the model. 
In contrast, later layers encode highly semantic representations, making them less transferable across variations in illumination or sensor parameters. 
As a result, source-affinity scores from these layers provide less reliable cues about their affinity to the source distribution.

\begin{table}[h]
    \centering
    \caption{\textbf{Ablation on layer depth for source-affinity score.} }
    \label{tab:layer_ablation}
    \fontsize{9pt}{10pt}\selectfont
    \begin{tabular}{c|cc|c}
        \toprule
        \textbf{Layer} & \textbf{IN-ES} & \textbf{IN-ES-Diverse} & \textbf{Latency (s)}\\
        \midrule
        1 & 87.65 & 67.25 & 1.22\\
        \textbf{1-3 (Ours)} & \textbf{87.85} & \textbf{67.28} & \textbf{1.32}\\
        1-12 & 82.65 & 18.22 & 1.40 \\
        10-12 & 82.55 & 18.85 & 1.35 \\
        12 & 82.75 & 20.02 & 1.40 \\
        \bottomrule
    \end{tabular}
\end{table}



\section{Controlled Comparison of Physical View Selection under Fixed $k$}
Table~\ref{tab:same_num_physical_views} presents a controlled comparison in which the number of selected physical views ($k$) and the aggregation strategy over filtered digital view predictions are fixed, while only the physical view selection criterion is varied. This setting allows us to isolate the effects of the two key components of \ours: (i) using multiple physical views as a new augmentation axis, and (ii) the proposed source affinity criterion.

Regardless of whether $k=1$ or $k=5$, the source affinity score of \ours~consistently outperforms the confidence score used in Lens~\cite{baek2025adaptive}, as it does not rely on physical views that the model may judge overconfidently. Instead, it selects physical views that are more aligned with the distribution learned by the model and therefore better understood by it, leading to higher performance.

In addition, Lens~\cite{baek2025adaptive} also benefits from multi-view aggregation, confirming the augmentation effect of using multiple physical views and further supporting the necessity of multi-physical view modeling, as emphasized in our approach. 

\begin{table}[t]
    \centering
    \caption{CLIP-ViT-B-16 initialized with MaPLe~\cite{khattak2023maple} weights}
    \label{tab:same_num_physical_views}
     \begin{tabular}{c|c|cc}
            \toprule
            \makecell{\# of \\ physical views} & \textbf{Method} & \textbf{IN-ES} & \makecell{\textbf{IN-ES} \\ \textbf{-Diverse}} \\
            \midrule
            \midrule
            \multirow[c]{2}{*}{\makecell{Single-view \\ ($k=1$)}} 
            & Lens + hard voting    & 84.30             & 65.05 \\
            \cmidrule{2-4}
            & \textbf{\ours (ours)}            & \textbf{85.45}    & \textbf{65.33} \\
            \midrule
            \multirow[c]{2}{*}{\makecell{Multi-view \\ ($k=5$)}} 
            & Lens + hard voting    & 86.95             & 66.78 \\
            \cmidrule{2-4}
            & \textbf{\ours (ours)}            & \textbf{87.85}    & \textbf{67.28} \\
            \bottomrule
        \end{tabular}
\end{table}

\section{Runtime latency}
\label{appendix:latency}
\begin{figure}[h]
    \centering
    \includegraphics[width=0.7\linewidth]{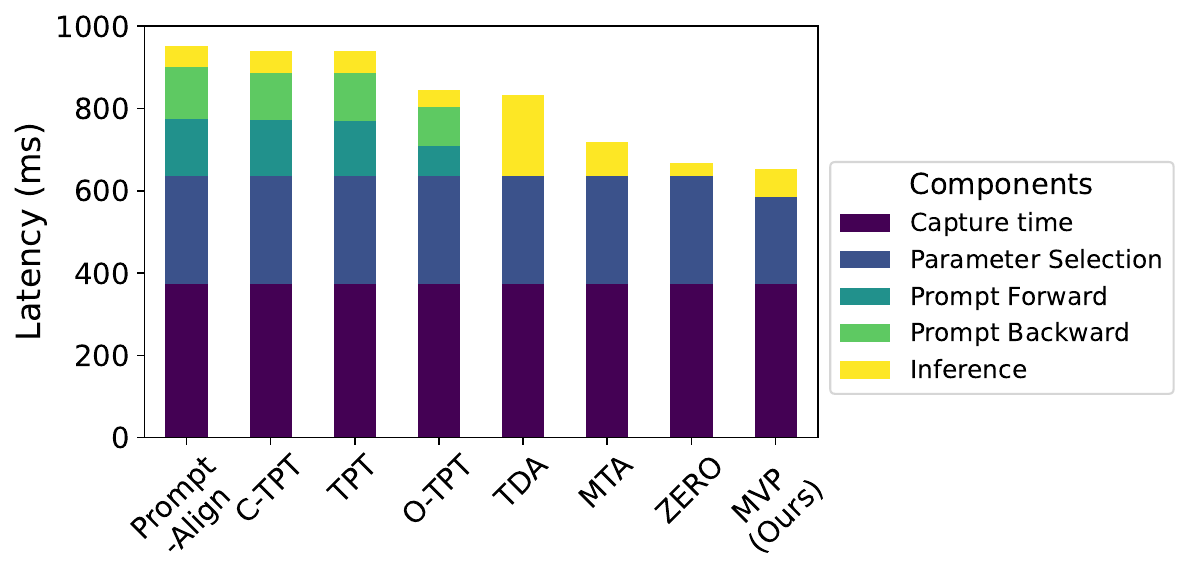}
    \caption{Total latency (ms) of adaptation and inference}
    \label{fig:supp_latency}
\end{figure}

Fig~\ref{fig:supp_latency} presents the runtime latency breakdown for baseline TTA methods and \ours~on ImageNet-ES-Diverse. All methods adopt CSA2 ($M$=6) to balance between capture cost and overall latency. The \textit{Parameter Selection} stage corresponds to selecting the most confident physical view in Lens and selecting the top-5 physical views with the highest source-affinity scores in \ours. Prompt-based TTA methods incur additional forward and backward passes to update prompt vectors during inference, while training-free TTAs and \ours~do not require such gradient-based operations. \textit{Inference} stage denotes the end-to-end forward pass that produces the final prediction for a single test sample.
Latency for parameter selection is averaged across all 27 parameter configurations, while latency for the remaining procedures is computed by measuring per-sample runtime over 1,000 test samples and reporting the average.
The total latency is illustrated in Fig~\ref{fig:latency}. 

\section{CSA details}
\label{appendix:csa}
CSA (Candidate Selection Algorithm) is a pre-selection strategy designed to reduce redundant capture time before performing sensor control~\cite{baek2025adaptive}. It provides three approaches for selecting $M$ candidate parameter configurations from the full set of available sensor-control options.

CSA1 performs uniform random sampling across all parameter configurations.

CSA2 adopts a grid-based random selection strategy to ensure even coverage of the parameter space. The full 3D (ISO, shutter speed, aperture) parameter space is partitioned into several grids, and one candidate is sampled from each grid until $M$ selections are made. Since parameters within the same grid are close in value, the resulting captures tend to exhibit similar image qualities. The number of grids is pre-defined as $1^3$ for $M=6$ in~\cite{baek2025adaptive}, but for fairer exposure variability, we adjust this to $2^3$ grids in our evaluation. 

CSA3 selects $M$ parameter configurations with the lowest capture cost, determined primarily by shutter speed. When multiple configurations have identical costs, the algorithm breaks ties randomly.

Table~\ref{tab:appendix_csa1},~\ref{tab:appendix_csa2}, and~\ref{tab:appendix_csa3} present extended CSA results, including those from Table~\ref{tab:csa_results}, evaluated across various values of $M$. For all experiments, we report the average performance over three runs with different random seeds. CSA3 places the strongest emphasis on efficiency among the three strategies, which leads to severe performance degradation especially under low-illumination conditions. Consequently, its accuracy drops sharply as $M$ decreases, and we therefore evaluate CSA3 only for $M=21$ and $M=18$.

\ours~consistently achieves the highest accuracy across all evaluated values of $M$.








\section{Limitations and future work} 
\label{appendix:limitations}
While \ours~operates in a fully forward-only manner and demonstrates strong robustness, it still requires more than $k$ physical captures per scene as well as pre-computed source-domain statistics. Future work includes extending the framework beyond static scenes and the restricted parameter space considered in this work—toward dynamic environments and broader regions of the exposure parameter manifold. Moreover, applying the proposed physical-prompting paradigm to tasks beyond classification (e.g., detection, retrieval, or semantic segmentation) represents a promising direction.

\newpage

\begin{table*}[]
    \centering
    \caption{Detailed results on ViT-B/16 backbone \textit{without} MaPLe}
    \label{tab:vit_no_maple}
    \begin{tabular}{c|ccc|ccccccc}
        \toprule
        \multicolumn{11}{c}{\textbf{Auto Exposure}} \\
        \midrule
        \multirow{2}{*}{\textbf{Method}} & \multicolumn{3}{c|}{{ImageNet-ES}} & \multicolumn{7}{c}{{ImageNet-ES-Diverse}} \\
        \cmidrule(lr){2-4} \cmidrule(lr){5-11}
        & {L1} & {L5} & {Average} & {L1} & {L2} & {L3} & {L4} & {L6} & {L7} & {Average} \\
        \midrule
        {CLIP} & 50.92 & 47.04 & 48.98 & 38.50 & 37.88 & 39.38 & 39.02 & 36.20 & 34.92 & 37.65 \\
        {TPT} & 54.65 & 50.44 & 52.56 & 40.80 & 40.72 & 42.04 & 40.74 & 38.32 & 36.96 & 39.93  \\
        {C-TPT} & 51.72 & 47.88 & 49.80 & 38.86 & 38.56 & 40.26 & 39.12 & 37.02 & 34.76 & 38.10 \\
        {O-TPT} & 51.06 & 46.84 & 48.95 & 38.30 & 37.40 & 39.02 & 38.08 & 35.82 & 33.92 & 37.09\\
        {MTA} & 54.42 & 50.60 & 52.51 & 40.66 & 39.52 & 40.82 & 40.38 & 38.22 & 36.78 & 39.40 \\
        {TDA} & 54.44 & 50.08 & 52.26 & 38.74 & 38.40 & 39.88 & 39.42 & 36.24 & 35.46 & 38.02 \\
        {ZERO} & 56.38 & 51.38 & 53.88 & 39.80 & 39.60 & 40.10 & 39.48 & 37.60 & 35.44 & 38.67 \\
        \midrule
        \multicolumn{11}{c}{\textbf{with Lens}} \\
        \midrule
        \multirow{2}{*}{\textbf{Method}} & \multicolumn{3}{c|}{{ImageNet-ES}} & \multicolumn{7}{c}{{ImageNet-ES-Diverse}} \\
        \cmidrule(lr){2-4} \cmidrule(lr){5-11}
        & {L1} & {L5} & {Average} & {L1} & {L2} & {L3} & {L4} & {L6} & {L7} & {Average} \\
        \midrule
        {CLIP+Lens} & 82.70 & 82.20 & 82.45 & 63.30 & 61.40 & 61.70 & 60.60 & 61.00 & 58.80 & 61.13 \\
        {TPT} & 82.80 & 82.00 & 82.40 & 63.70 & 62.40 & 62.20 & 61.30 & 61.20 & 59.40 & 61.70 \\
        {C-TPT} & 82.80 & 82.20 & 82.50 & 63.60 & 61.90 & 62.50 & 60.60 & 61.40 & 59.20 & 61.53 \\
        {O-TPT} & 82.70 & 81.80 & 82.25 & 63.10 & 61.30 & 62.10 & 60.60 & 60.80 & 59.30 & 61.20 \\
        {MTA} & 82.60 & 82.10 & 82.35 & 63.70 & 61.90 & 62.40 & 60.80 & 60.70 & 59.30 & 61.47 \\
        {TDA} & 82.70 & 82.40 & 82.55 & 62.20 & 61.10 & 61.20 & 60.70 & 60.20 & 58.70 & 60.58 \\
        {ZERO} & 82.90 & 82.40 & 82.65 & 63.60 & 61.20 & 62.10 & 61.70 & 59.90 & 58.80 & 61.22 \\
        \midrule
        \textbf{\ours~(ours)} & \textbf{85.30} & \textbf{84.30} & \textbf{84.80} & \textbf{65.30} & \textbf{64.80} & \textbf{65.50} & \textbf{64.20} & \textbf{63.40} & \textbf{62.70} & \textbf{64.32} \\
        \bottomrule
    \end{tabular}
\end{table*}


\begin{table*}[]
    \centering
    \caption{Detailed results on ResNet50 backbone}
    \label{tab:rn50}
    \begin{tabular}{c|ccc|ccccccc}
        \toprule
        \multicolumn{11}{c}{\textbf{Auto Exposure}} \\
        \midrule
        \multirow{2}{*}{\textbf{Method}} & \multicolumn{3}{c|}{{ImageNet-ES}} & \multicolumn{7}{c}{{ImageNet-ES-Diverse}} \\
        \cmidrule(lr){2-4} \cmidrule(lr){5-11}
        & {L1} & {L5} & {Average} & {L1} & {L2} & {L3} & {L4} & {L6} & {L7} & {Average} \\
        \midrule
        {CLIP} & 35.88 & 31.96 & 33.92 & 17.44 & 17.70 & 18.32 & 17.66 & 16.26 & 15.52 & 17.15 \\
        {TPT} & 39.50 & 36.18 & 37.84 & 21.08 & 21.60 & 21.82 & 21.36 & 20.00 & 18.70 & 20.76 \\
        {C-TPT} & 37.62 & 33.12 & 35.37 & 17.74 & 18.06 & 18.74 & 18.06 & 16.56 & 15.74 & 17.48 \\
        {O-TPT} & 36.08 & 31.68 & 33.97 & 18.34 & 18.24 & 19.02 & 18.52 & 17.26 & 16.46 & 17.97 \\
        {MTA} & 38.52 & 34.90 & 36.71 & 16.76 & 16.52 & 17.36 & 16.34 & 15.20 & 14.22 & 16.15 \\
        {TDA} & 38.62 & 35.16 & 36.89 & 16.16 & 16.26 & 16.86 & 16.32 & 14.24 & 13.60 & 15.57 \\
        {ZERO} & 39.34 & 35.82 & 37.58 & 15.82 & 15.42 & 15.84 & 15.96 & 14.22 & 13.10 & 15.06 \\
        \midrule
        \multicolumn{11}{c}{\textbf{with Lens}} \\
        \midrule
        \multirow{2}{*}{\textbf{Method}} & \multicolumn{3}{c|}{{ImageNet-ES}} & \multicolumn{7}{c}{{ImageNet-ES-Diverse}} \\
        \cmidrule(lr){2-4} \cmidrule(lr){5-11}
        & {L1} & {L5} & {Average} & {L1} & {L2} & {L3} & {L4} & {L6} & {L7} & {Average} \\
        \midrule
        {CLIP+Lens} & 73.60 & 73.30 & 73.45 & 40.90 & 41.50 & 39.90 & 38.70 & 38.60 & 37.20 & 39.47 \\
        {TPT} & 73.90 & 73.70 & 73.80 & 41.40 & 41.50 & 40.40 & 39.60 & 39.30 & 37.40 & 39.93 \\
        {C-TPT} & 73.70 & 73.40 & 73.55 & 41.30 & 41.20 & 40.20 & 39.60 & 39.30 & 37.60 & 39.87 \\
        {O-TPT} & 73.70 & 73.40 & 73.55 & 41.40 & 41.60 & 40.30 & 39.40 & 39.00 & 37.10 & 39.80 \\
        {MTA} & 74.00 & 73.70 & 73.85 & 41.20 & 41.00 & 39.90 & 39.00 & 38.40 & 37.40 & 39.48 \\
        {TDA} & 72.50 & 73.40 & 72.95 & 40.70 & 39.60 & 39.80 & 38.60 & 37.80 & 36.00 & 38.75 \\
        {ZERO} & 73.50 & 73.90 & 73.70 & 40.20 & 39.80 & 40.00 & 37.60 & 38.20 & 35.20 & 38.50 \\
        \midrule
        \textbf{\ours~(ours)} & \textbf{76.00} & \textbf{74.80} & \textbf{75.40} & \textbf{43.20} & \textbf{41.70} & \textbf{41.50} & \textbf{38.80} & \textbf{39.10} & \textbf{36.50} & \textbf{40.13} \\
        \bottomrule
    \end{tabular}
\end{table*}



\begin{table*}[t]
    \centering
    \caption{Detailed results of main evaluation}
    \label{tab:main_detail}
    \begin{tabular}{c|ccc|ccccccc}
        \toprule
        \multicolumn{11}{c}{\textbf{Auto Exposure}} \\
        \midrule
        \multirow{2}{*}{\textbf{Method}} & \multicolumn{3}{c|}{{ImageNet-ES}} & \multicolumn{7}{c}{{ImageNet-ES-Diverse}} \\
        \cmidrule(lr){2-4} \cmidrule(lr){5-11}
        & {L1} & {L5} & {Average} & {L1} & {L2} & {L3} & {L4} & {L6} & {L7} & {Average} \\
        \midrule
        CLIP & 50.92 & 47.04 & 48.98 & 38.50 & 37.88 & 39.38 & 39.02 & 36.20 & 34.92 & 37.65 \\
        CoOp & 54.48 & 49.16 & 51.82 & 41.36 & 40.78 & 42.20 & 41.72 & 39.36 & 37.38 & 40.47 \\
        MaPle & 54.98 & 50.16 & 52.57 & 40.56 & 39.76 & 41.24 & 40.24 & 38.48 & 36.52 & 39.47 \\
        TPT & 58.64 & 52.68 & 55.66 & 42.86 & 41.34 & 43.00 & 41.76 & 39.62 & 38.60 & 41.20 \\
        PromptAlign & 58.20 & 52.70 & 55.45 & 42.66 & 41.40 & 43.70 & 42.32 & 39.90 & 39.06 & 41.51 \\
        C-TPT & 56.24 & 50.86 & 53.55 & 41.24 & 40.16 & 41.46 & 40.22 & 39.02 & 37.18 & 39.88 \\
        O-TPT & 55.12 & 50.28 & 52.70 & 40.38 & 39.24 & 40.56 & 39.98 & 38.26 & 36.42 & 39.14 \\
        MTA & 59.30 & 53.82 & 56.56 & 43.14 & 42.04 & 43.60 & 42.10 & 40.20 & 39.10 & 41.70 \\
        TDA & 61.16 & 55.18 & 58.17 & 41.96 & 41.06 & 42.64 & 41.42 & 39.30 & 38.30 & 40.78 \\
        ZERO & 59.92 & 54.18 & 57.05 & 40.90 & 40.36 & 41.82 & 40.88 & 38.34 & 37.16 & 39.91 \\
        \midrule
        \multicolumn{11}{c}{\textbf{Auto Exposure + Photo. Aug.}} \\
        \midrule
        \multirow{2}{*}{\textbf{Method}} & \multicolumn{3}{c|}{{ImageNet-ES}} & \multicolumn{7}{c}{{ImageNet-ES-Diverse}} \\
        \cmidrule(lr){2-4} \cmidrule(lr){5-11}
        & {L1} & {L5} & {Average} & {L1} & {L2} & {L3} & {L4} & {L6} & {L7} & {Average} \\
        \midrule
        {TPT} & 52.12 & 47.20 & 49.66 & 42.80 & 40.88 & 42.96 & 41.78 & 38.92 & 37.60 & 40.82 \\
        {PromptAlign} & 53.18 & 47.36 & 50.27 & 42.98 & 41.06 & 42.94 & 41.60 & 39.82 & 37.86 & 41.04 \\
        {C-TPT} & 48.52 & 43.32 & 45.92 & 40.94 & 39.62 & 40.64 & 40.16 & 37.54 & 35.64 & 39.09 \\
        {O-TPT} & 47.00 & 42.12 & 44.56 & 40.36 & 38.90 & 39.94 & 39.30 & 36.88 & 35.42 & 38.47 \\
        {MTA} & 57.22 & 52.06 & 54.64 & 42.84 & 41.66 & 43.56 & 42.14 & 39.52 & 38.26 & 41.33 \\
        {TDA} & 60.44 & 54.14 & 57.29 & 42.12 & 41.06 & 42.58 & 41.52 & 39.50 & 38.20 & 40.83 \\
        {ZERO} & 59.56 & 53.98 & 56.77 & 41.20 & 39.62 & 41.74 & 40.26 & 38.52 & 37.46 & 39.80 \\
        \midrule
        \multicolumn{11}{c}{\textbf{with Lens}} \\
        \midrule
        \multirow{2}{*}{\textbf{Method}} & \multicolumn{3}{c|}{{ImageNet-ES}} & \multicolumn{7}{c}{{ImageNet-ES-Diverse}} \\
        \cmidrule(lr){2-4} \cmidrule(lr){5-11}
        & {L1} & {L5} & {Average} & {L1} & {L2} & {L3} & {L4} & {L6} & {L7} & {Average} \\
        \midrule
        {CLIP+Lens} & 82.70 & 82.20 & 82.45 & 63.30 & 61.40 & 61.70 & 60.60 & 61.00 & 58.80 & 61.13 \\
        {CoOp} & 84.80 & {85.00} & {84.90} & {68.20} & {66.10} & 65.70 & 64.20 & {64.40} & 62.30 & {65.15} \\
        {MaPle} & 85.10 & 83.70 & 84.40 & 65.90 & 64.80 & 66.20 & 64.90 & 63.60 & 62.70 & 64.68 \\
        {TPT} & 85.10 & 83.80 & 84.45 & 66.50 & 65.10 & 66.30 & 65.00 & 64.10 & 62.40 & 64.90 \\
        {PromptAlign} & 85.70 & 83.80 & 84.75 & 66.90 & 65.30 & 65.70 & 65.50 & 64.00 & 62.00 & 64.90 \\
        {C-TPT} & 85.30 & 83.80 & 84.55 & 66.00 & 64.90 & 66.10 & 64.70 & 63.90 & 62.70 & 64.72 \\
        {O-TPT} & 85.30 & 83.70 & 84.50 & 66.00 & 64.80 & 66.10 & 64.50 & 63.80 & 62.60 & 64.63 \\
        {MTA} & 85.40 & 84.00 & 84.70 & 66.20 & 65.70 & {66.50} & 65.20 & 64.20 & 62.70 & 65.08 \\
        {TDA} & {85.80} & 83.80 & 84.80 & 65.90 & 64.40 & 66.30 & {66.00} & 63.60 & {62.90} & 64.85 \\
        {ZERO} & 85.20 & 84.40 & 84.80 & 66.00 & 65.20 & 66.00 & 65.10 & 63.30 & 62.40 & 64.67 \\
        \midrule
        \multicolumn{11}{c}{\textbf{Ours}} \\
        \midrule
        \textbf{\ours~(ours)} &  87.50 & \textbf{88.20} & \textbf{87.85} & \textbf{69.30} & \textbf{67.50} & \textbf{69.20} & \textbf{66.70} & \textbf{66.00} & \textbf{65.00} & \textbf{67.28} \\
        \bottomrule
    \end{tabular}
\end{table*}

\begin{table*}[t]
    \centering
    \caption{\textbf{Results with CSA1 across different $M$.}
    }
    \label{tab:appendix_csa1}
    \begin{tabular}{c|c|c|c|c|c|c|c|c|c|c}
        \toprule
        \multirow{4}{*}{\textbf{Method}} & \multicolumn{2}{c|}{\makecell{\textbf{$M=18$}}} & \multicolumn{2}{c|}{\makecell{\textbf{$M=15$}}} & \multicolumn{2}{c|}{\makecell{\textbf{$M=12$}}} & \multicolumn{2}{c|}{\makecell{\textbf{$M=9$}}} & \multicolumn{2}{c}{\makecell{\textbf{$M=6$}}} \\
        \cline{2-11} 

        & \multirow[c]{2}{*}{\footnotesize\textbf{IN-ES}} 
        & \multirow[c]{2}{*}{\footnotesize\textbf{\makecell{IN-ES\\-Diverse}}}
        & \multirow[c]{2}{*}{\footnotesize\textbf{IN-ES}}
        & \multirow[c]{2}{*}{\footnotesize\textbf{\makecell{IN-ES\\-Diverse}}}
        & \multirow[c]{2}{*}{\footnotesize\textbf{IN-ES}}
        & \multirow[c]{2}{*}{\footnotesize\textbf{\makecell{IN-ES\\-Diverse}}}
        & \multirow[c]{2}{*}{\footnotesize\textbf{IN-ES}}
        & \multirow[c]{2}{*}{\footnotesize\textbf{\makecell{IN-ES\\-Diverse}}}
        & \multirow[c]{2}{*}{\footnotesize\textbf{IN-ES}}
        & \multirow[c]{2}{*}{\footnotesize\textbf{\makecell{IN-ES\\-Diverse}}} \\
        & & & & & & & & & & \\
        \midrule
        Lens & 84.73 & 63.53 & 84.82 & 63.52 & 84.75 & 61.79 & 84.48 & 60.48 & 83.33 & 55.40 \\
        \midrule
        + TPT & 84.82 & 63.94 & 84.98 & 63.86 & 84.88 & 62.22 & 84.90 & 61.03 & 83.77 & 56.33 \\
        + PromptAlign & 84.98 & 63.74 & 85.25 & 63.82 & 84.95 & 62.03 & 84.82 & 61.01 & 84.03 & 56.24  \\
        + C-TPT & 84.88 & 63.56 & 84.83 & 63.56 & 84.85 & 61.90 & 84.83 & 60.66 & 83.65 & 53.52  \\
        + O-TPT & 84.75 & 63.39 & 84.75 & 63.41 & 84.77 & 61.66 & 84.57 & 60.34 & 83.43 & 51.06 \\
        + MTA & 85.18 & 64.09 & 85.23 & 64.09 & 85.37 & 62.44 & 85.12 & 61.17 & 84.02 & 56.51 \\
        + TDA & 85.22 & 63.81 & 85.37 & 63.53 & 85.30 & 61.71 & 85.12 & 60.64 & 84.23 & 56.02  \\
        + ZERO & 85.43 & 63.50 & 85.75 & 63.38 & 85.43 & 61.49 & 85.23 & 60.03 & 84.60 & 55.59  \\
        \midrule
        \textbf{\ours~(ours)} & \textbf{87.82} & \textbf{66.02} & \textbf{87.57} & \textbf{65.97} & \textbf{87.27} & \textbf{63.79} & \textbf{86.40} & \textbf{62.27} & \textbf{85.40} & \textbf{57.49}  \\
        \bottomrule
    \end{tabular}
    \vspace{-2ex}
\end{table*}

\begin{table*}[t]
    \centering
    \caption{\textbf{Results with CSA2 across different $M$.}
    }
    \label{tab:appendix_csa2}
    \begin{tabular}{c|c|c|c|c|c|c|c|c|c|c}
        \toprule
        \multirow{4}{*}{\textbf{Method}} & \multicolumn{2}{c|}{\makecell{\textbf{$M=18$}}} & \multicolumn{2}{c|}{\makecell{\textbf{$M=15$}}} & \multicolumn{2}{c|}{\makecell{\textbf{$M=12$}}} & \multicolumn{2}{c|}{\makecell{\textbf{$M=9$}}} & \multicolumn{2}{c}{\makecell{\textbf{$M=6$}}} \\
        \cline{2-11} 

        & \multirow[c]{2}{*}{\footnotesize\textbf{IN-ES}} 
        & \multirow[c]{2}{*}{\footnotesize\textbf{\makecell{IN-ES\\-Diverse}}}
        & \multirow[c]{2}{*}{\footnotesize\textbf{IN-ES}}
        & \multirow[c]{2}{*}{\footnotesize\textbf{\makecell{IN-ES\\-Diverse}}}
        & \multirow[c]{2}{*}{\footnotesize\textbf{IN-ES}}
        & \multirow[c]{2}{*}{\footnotesize\textbf{\makecell{IN-ES\\-Diverse}}}
        & \multirow[c]{2}{*}{\footnotesize\textbf{IN-ES}}
        & \multirow[c]{2}{*}{\footnotesize\textbf{\makecell{IN-ES\\-Diverse}}}
        & \multirow[c]{2}{*}{\footnotesize\textbf{IN-ES}}
        & \multirow[c]{2}{*}{\footnotesize\textbf{\makecell{IN-ES\\-Diverse}}} \\
        & & & & & & & & & & \\
        \midrule
        Lens & 84.62 & 63.31 & 84.52 & 63.21 & 84.28 & 63.29 & 84.43 & 61.10 & 84.38 & 61.76  \\
        \midrule
        + TPT & 84.52 & 63.78 & 84.47 & 63.66 & 84.28 & 63.78 & 84.48 & 61.54 & 84.55 & 62.49 \\
        + PromptAlign & 84.65 & 63.48 & 84.52 & 63.73 & 84.25 & 63.67 & 84.60 & 61.53 & 84.32 & 62.56  \\
        + C-TPT & 84.72 & 63.33 & 84.65 & 63.31 & 84.47 & 63.41 & 84.60 & 60.97 & 84.68 & 61.83 \\
        + O-TPT & 84.62 & 63.29 & 84.50 & 63.22 & 84.37 & 63.27 & 84.53 & 60.48 & 84.42 & 61.27 \\
        + MTA & 84.98 & 63.89 & 84.87 & 63.89 & 84.88 & 64.04 & 84.95 & 61.80 & 85.10 & 62.74 \\
        + TDA & 84.83 & 63.52 & 85.00 & 63.23 & 84.82 & 63.36 & 84.38 & 61.26 & 84.70 & 62.08  \\
        + ZERO & 85.28 & 63.19 & 85.10 & 63.06 & 84.70 & 63.16 & 84.75 & 60.86 & 84.60 & 61.96 \\
        \midrule
        \textbf{\ours~(ours)} & \textbf{87.68} & \textbf{65.68} & \textbf{87.48} & \textbf{65.44} & \textbf{86.90} & \textbf{65.47} & \textbf{86.92} & \textbf{63.09} & \textbf{86.65} & \textbf{63.79}  \\
        \bottomrule
    \end{tabular}
    \vspace{-2ex}
\end{table*}

\begin{table*}[t]
    \centering
    \caption{\textbf{Results with CSA3 across different $M$.}
    }
    \label{tab:appendix_csa3}
    \begin{tabular}{c|c|c|c|c}
        \toprule
        \multirow{4}{*}{\textbf{Method}} 
        & \multicolumn{2}{c|}{\makecell{\textbf{$M=21$}}} 
        & \multicolumn{2}{c}{\makecell{\textbf{$M=18$}}}\\
        \cline{2-5} 

        & \multirow[c]{2}{*}{\footnotesize\textbf{IN-ES}} 
        & \multirow[c]{2}{*}{\footnotesize\textbf{\makecell{IN-ES\\-Diverse}}}
        & \multirow[c]{2}{*}{\footnotesize\textbf{IN-ES}}
        & \multirow[c]{2}{*}{\footnotesize\textbf{\makecell{IN-ES\\-Diverse}}} \\
        & & & & \\
        \midrule
        Lens &  84.98 & 61.67 & 85.00 & 54.38 \\
        \midrule
        + TPT & 85.05 & 62.29 & 85.15 & 55.32 \\
        + PromptAlign & 85.25 & 62.37 & 85.40 & 55.40  \\
        + C-TPT & 85.10 & 61.91 & 85.10 & 54.58  \\
        + O-TPT & 85.03 & 61.77 & 85.05 & 54.40 \\
        + MTA & 85.33 & 62.62 & 85.30 & 55.30  \\
        + TDA & 85.28 & 62.33 & 85.05 & 54.93 \\
        + ZERO & 85.62 & 61.69 & 85.90 & 54.28  \\
        \midrule
        \textbf{\ours~(ours)} & \textbf{87.87} & \textbf{64.41} & \textbf{87.70} & \textbf{55.92} \\
        \bottomrule
    \end{tabular}
    \vspace{-2ex}
\end{table*}

\end{document}